\documentclass[10pt,twocolumn,letterpaper]{article}

\usepackage{graphicx}
\usepackage{amsmath}
\usepackage{amssymb}
\usepackage{booktabs}
\usepackage{pgfplots}
\usepgfplotslibrary{fillbetween}
\usetikzlibrary{arrows}
\usepackage{tikz}
\usepackage{url}
\usepackage{xpatch}
\usepackage{xcolor}
\usepackage{listings}
\usepackage{realboxes}
\usepackage{adjustbox}
\usepackage{caption}
\usepackage[accsupp]{axessibility}
\usepackage[format=plain,labelformat=simple,labelsep=period,font=small,compatibility=false]{caption}
\usepackage[font=footnotesize,skip=3pt,subrefformat=parens]{subcaption}
\usepackage{cite}
\usepackage{xspace}
\usepackage[english]{babel}

\definecolor{mygray}{rgb}{0.8,0.8,0.8}
\makeatletter
\DeclareRobustCommand\onedot{\futurelet\@let@token\@onedot}
\def\@onedot{\ifx\@let@token.\else.\null\fi\xspace}
\def\eg{\emph{e.g}\onedot}

\def\etal{\emph{et al}\onedot}
\makeatother
\makeatletter
\xpretocmd\lstinline{\Colorbox{mygray}\bgroup\appto\lst@DeInit{\egroup}}{}{}
\makeatother

\usepackage[pagebackref,breaklinks,colorlinks, bookmarks=false]{hyperref}

\usepackage[capitalize]{cleveref}
\crefname{section}{Sec.}{Secs.}
\Crefname{section}{Section}{Sections}
\Crefname{table}{Table}{Tables}
\crefname{table}{Tab.}{Tabs.}

\begin{document}

\title{Robustness of Visual Explanations to Common Data Augmentation Methods}
\author{~~~Lenka Tětková ~~~~~~~~~ Lars Kai Hansen \\
~~~~{\tt\small lenhy@dtu.dk} ~~~~~~~~~~~~
  {\tt\small lkai@dtu.dk} ~~~~~~\\
  Technical University of Denmark \\
Department of Applied Mathematics and Computer Science \\ 
Richard Petersens Plads, 321, 2800 Kgs. Lyngby, Denmark}
\date{}
\maketitle

\begin{abstract}
As the use of deep neural networks continues to grow, understanding their behaviour has become more crucial than ever. Post-hoc explainability methods are a potential solution, but their reliability is being called into question. Our research investigates the response of post-hoc visual explanations to naturally occurring transformations, often referred to as augmentations. We anticipate explanations to be invariant under certain transformations, such as changes to the colour map while responding in an equivariant manner to transformations like translation, object scaling, and rotation. We have found remarkable differences in robustness depending on the type of transformation, with some explainability methods (such as LRP composites and Guided Backprop) being more stable than others. We also explore the role of training with data augmentation. We provide evidence that explanations are typically less robust to augmentation than classification performance, regardless of whether data augmentation is used in training or not.
\end{abstract}
\vspace{-3mm}
%
\section{Introduction}
\label{sec:intro}
Convolutional neural networks (CNNs) are commonly used in computer vision. However, CNNs are fragile to adversarial attacks \cite{goodfellow2014explaining}.  It has been shown  that explanation methods are fragile as well and that attackers can manipulate the explanations arbitrarily \cite{dombrowski2019explanations, ghorbani2019interpretation}.

To be trusted, explanations need to show common-sense behaviour. In this work, we investigate one such basic behaviour:
{\it If a transformation of an image does not change the target class, the explanation should assign importance to the same part of the object as in the untransformed image}\footnote{We do not consider cases where the transformation of an image would change the ground-truth label.}.
If the explainability method does not preserve the explanations of the perturbed images, we lose trust in it. We believe that it is even more concerning than adversarial attacks since perturbations such as \eg, object rotation, are omnipresent and happen spontaneously. 

In this work, we investigate how perturbations of an image influence visual post-hoc explanations. To understand the role of augmentation during training, we train CNNs from scratch on both augmented and non-augmented data. We examine the robustness of the models and compare the explanations. We pose the questions: Are visual explanations as robust to augmenting the input image as the predictions? Are there differences among various explainability methods and model architectures? Does training with augmented images improve the robustness? Which explainability methods are the best both in robustness to small augmentations and in faithfulness measured by the pixel-flipping test?
\paragraph{Related work}
The feasibility of adversarial attacks \cite{akhtar2018threat} is well-known. It has been shown \cite{dombrowski2019explanations, ghorbani2019interpretation} that explanation methods are fragile as well and that attackers can manipulate the explanations arbitrarily. In this paper, we focus on the fragility of the explanations in the case of more naturally occurring (often unintentional) disruptions.

Data augmentation techniques \cite{NIPS1996_81e5f81d, Cire_an_2010} have been used to improve the generalization of the image classifiers (\eg, \cite{wang2017effectiveness, shorten2019survey}). Rebuffi \etal \cite{rebuffi2021data} found that using data augmentations helps to improve the robustness against adversarial attacks. Very recent work by Won \etal \cite{Won_Bae_Kim_2023} found that data augmentation used under model training has an impact on model interpretability, however, they do not consider stability under test time augmentation as in the present work.

Wang and Wang \cite{wang2021selfinterpretable} built a model with transformation invariant interpretations. However, this self-interpretable model violates one of the desiderata for explanations \cite{Swartout_Moore_1993}: low construction overhead. We explore whether we could get similar robustness with available post-hoc explainability methods. Moreover, we broaden the set of considered transformations.

Although explainability is important for understanding neural networks, the existing methods differ in the quality of produced explanations and many saliency methods have been criticized (\eg, \cite{Adebayo_Gilmer_Muelly_Goodfellow_Hardt_Kim_2018, Kindermans_Hooker_Adebayo_Alber_Schutt_Dahne_Erhan_Kim_2017, Nie_Zhang_Patel_2020}). Therefore, metrics to evaluate the quality have been developed (\eg, \cite{Bach_Binder_Montavon_Klauschen_Muller_Samek_2015, Rieger_Hansen_2020, Chalasani_Chen_Chowdhury_Jha_Wu_2020}). Quantus \cite{hedstrom2023quantus} is a toolkit that collects many of those metrics. Our experiments shed further light on the stability of explainability methods.
%
\section{Methods}
\vspace{-1mm}
\paragraph{Augmentation methods}
Here we divide augmentation techniques into two groups: invariant and equivariant methods. For invariant techniques, the explanation of the augmented image should be the same as the explanation of the original image. In the case of the equivariant techniques, the explanation of the augmented image should be the same as the augmented explanation of the original image.

We chose three invariant (change of brightness, hue and saturation) and three equivariant techniques (rotation, translation and scaling). When using equivariant methods, the background of each image was padded with black pixels to match the original image format if necessary. In preliminary experiments, we studied the influence of various background padding methods and differences were negligible.

\begin{figure}
  \begin{subfigure}{\linewidth}
   \includegraphics[width=\linewidth]{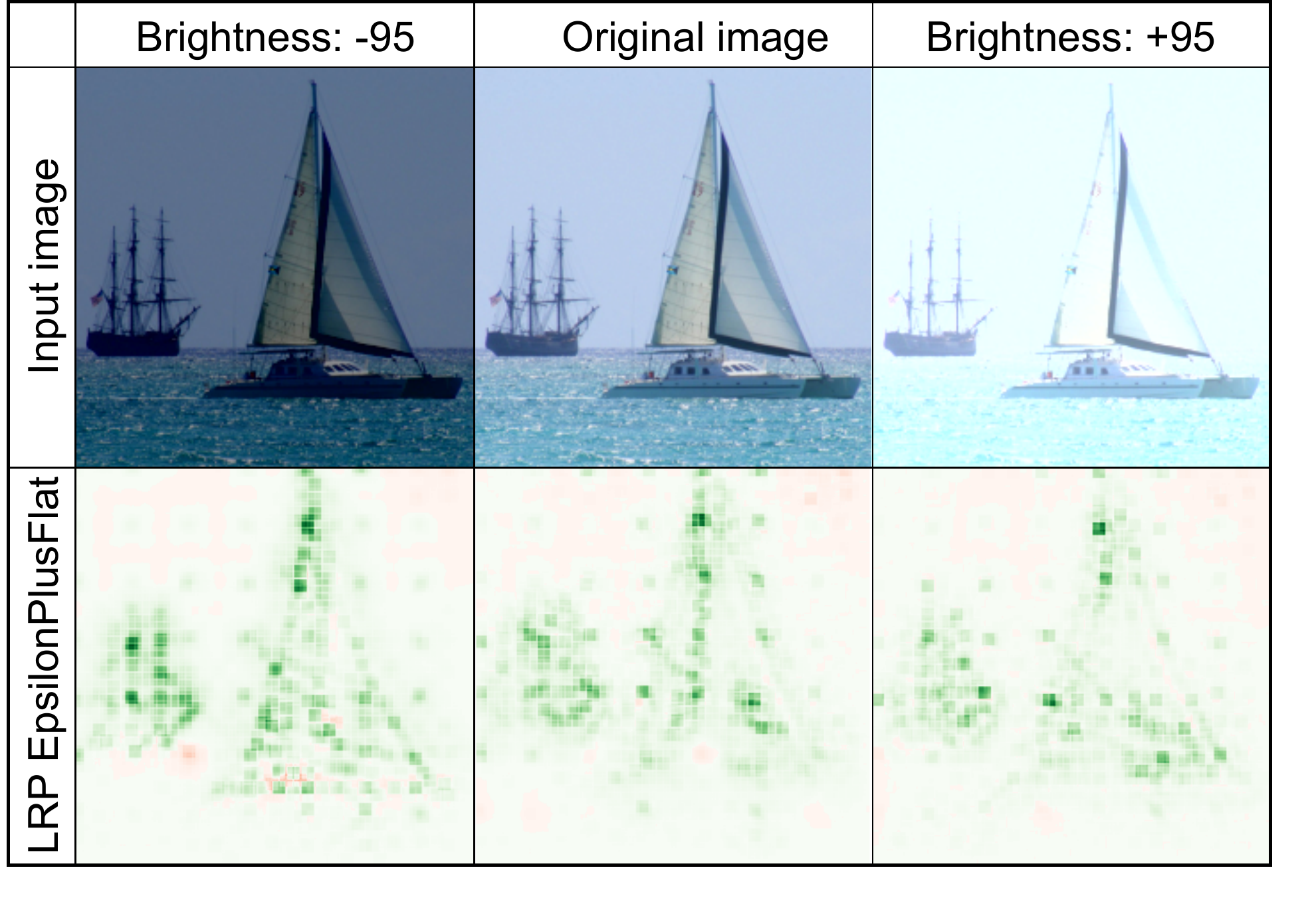}%
   \caption{Brightness}%
   \label{fig:aug_brightness}%
  \end{subfigure}\par
  \begin{subfigure}{0.97\linewidth}
   \includegraphics[width=\linewidth]{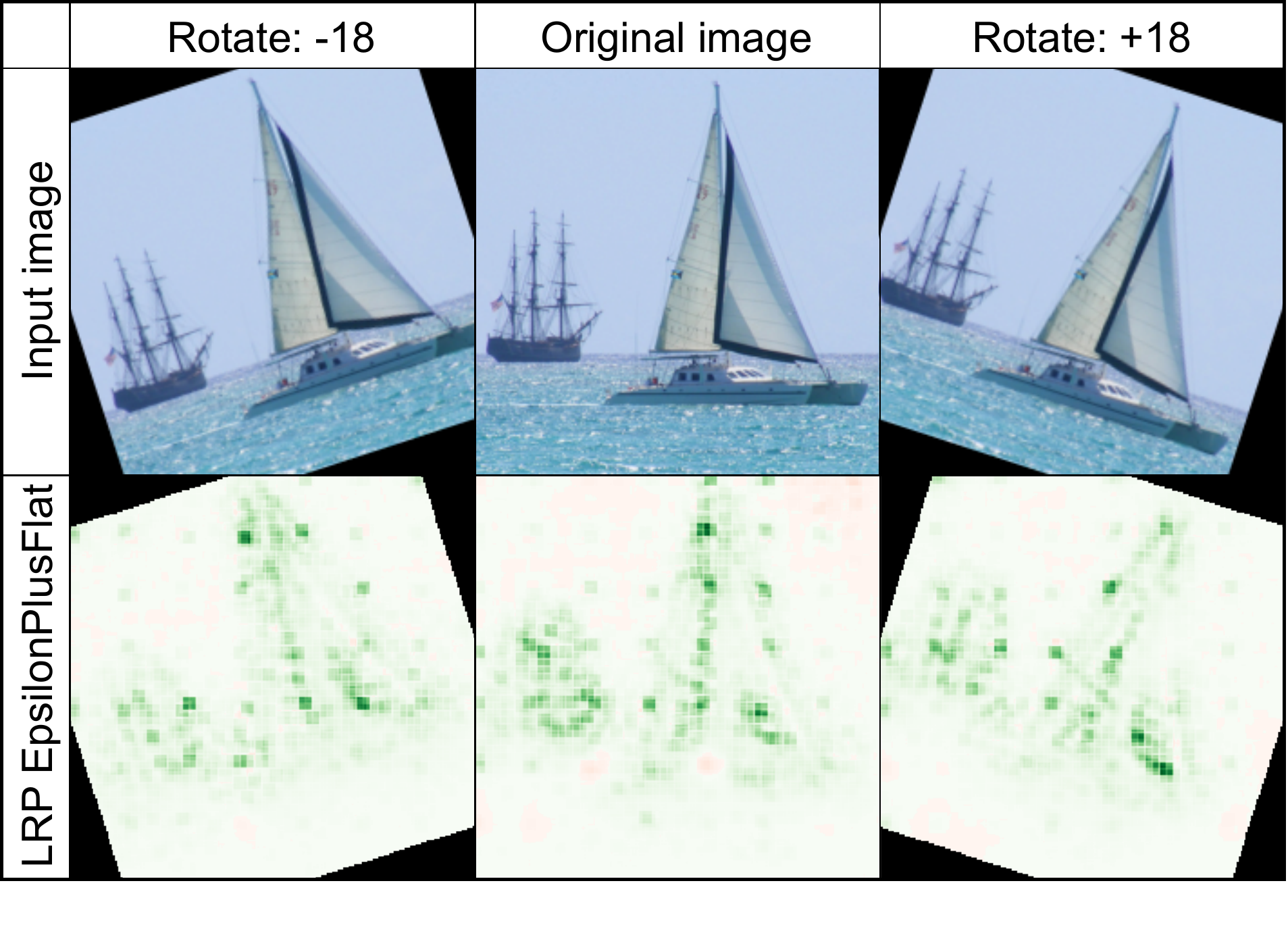}
   \caption{Rotate}
   \label{fig:aug_rotate}
   \vspace{-3mm}
  \end{subfigure}
   \caption{Examples of the augmented images and their explanations.}
   \label{fig:aug_examples}
   \vspace{-8mm}
\end{figure}

We used the library ImgAug \cite{imgaug} for augmenting images. For each method, we chose an interval of values so that classification performance was reduced by 10\%. A table showing the chosen intervals for each method can be found in \cref{tab:aug_intervals} and \cref{fig:aug_examples} displays one image augmented by values within these intervals for changing brightness and rotation. The figures for the rest of the methods are in \cref{appendix:aug_methods}.

\begin{figure}
  \centering
  \begin{subfigure}[h]{0.23\textwidth}
  \begin{adjustbox}{width=\linewidth}
    \begin{tikzpicture}
  
    \begin{axis}[
        xmin = -2, xmax = 2,
        ymin = 0, ymax = 1,
        xtick = {-1, 0, 1},
        xticklabels= {$M$, $0$, $N$},
        grid = both,
        major grid style = {lightgray},
        minor grid style = {lightgray!25},
        xlabel = {augmentation parameter},
        ylabel = {target class probability},
        label style={font=\fontsize{16}{0}\selectfont},]
     
    \addplot[
        domain = -3:3,
        samples = 50,
        smooth,
        thick,
        blue,
    ] {-1/5*(x^2)+0.6};

    \addplot[
        domain = -3:3,
        samples = 50,
        smooth,
        thick,
        dashed,
        blue,
        name path=A,
    ] {-1/5*(x^2)+1};
    
    \addplot[name path=B] coordinates {(-1,0)(1,0)};
    \addplot[->, red, thick] coordinates {(0,0.6) (0,1)};
    
    \addplot[color=green!40!black, ultra thick] coordinates {(-1, 0) (1, 0) (1, 1) (-1, 1) (-1, 0)};
    \addplot[yellow, fill opacity=0.5] fill between[of=A and B, soft clip={domain=-1:1}];
    
    \end{axis}
     
    \end{tikzpicture}
     \end{adjustbox}   
    \subcaption{Robustness score}
    \label{fig:Score-AOC}
  \end{subfigure}
  \hfill
  \begin{subfigure}[h]{0.23\textwidth}
    \begin{adjustbox}{width=\linewidth}
    \begin{tikzpicture}
 
    \begin{axis}[
        xmin = 0, xmax = 0.3,
        ymin = 0, ymax = 1,
        xtick distance = 0.1,
        grid = both,
        major grid style = {lightgray},
        minor grid style = {lightgray!25},
        ylabel = {$f(x_{augmented}/f(x_{orig})$},
        xlabel = {portion of the pixels flipped},
        label style={font=\fontsize{16}{0}\selectfont},]
     
    \addplot[
        domain = 0:1,
        samples = 50,
        smooth,
        thick,
        blue,
        name path=A,
    ] {1/(5*x^0.25+1)};

    \addplot[const plot, name path=B] coordinates {(0,1)(0.2,1)};
    
    \addplot[color=green!40!black, ultra thick] coordinates {(0, 0) (0.2, 0) (0.2, 1) (0, 1) (0,0)};
    \addplot[yellow, fill opacity=0.5] fill between[of=A and B, soft clip={domain=0:0.2}];
    
    \end{axis}
     
    \end{tikzpicture}
     \end{adjustbox}   
    \subcaption{Pixel flipping}
    \label{fig:PF-AOC}
  \end{subfigure}
  \caption{Visualization of the metrics defined in \cref{sec:methods_metrics}. In both cases, we compute the portion of the yellow part in the green rectangle.}
  \label{fig:metrics}
  \vspace{-5mm}
\end{figure}
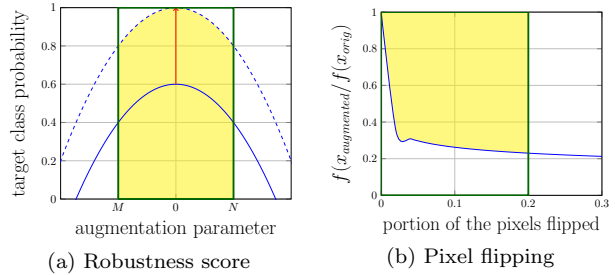

The experiments were performed on the ImageNet \cite{imagenet_cvpr09} dataset. For comparing explanations, 500 images across all ImageNet classes were randomly selected. Analyses were done on the correctly classified images. For every augmentation method and for each image, the interval of possible values of the augmentation method parameter was divided into equidistant units and augmented versions of the image were created, one for each of these values. Each image was passed through the networks to get the probability of the target class and we got explanations for post-hoc explainability methods. We computed the Pearson correlation between the explanations of the augmented images and the explanation of the original image (augmented explanation in the case of the equivariant methods) and top-1000 intersection (intersection of 1000 most important pixels in the explanation). We compared only the area of the original image -- hence, in equivariant methods, we computed the correlation and top-1000 intersection only on the parts that were present in both the original and the augmented image and mask the rest. 
\paragraph{Metrics}\label{sec:methods_metrics}
We can plot the probability of the target class and all its augmented versions with the augmentation parameter on the x-axis and the probability on the y-axis.  We call this relation a \textit{probability curve}. In the same way, we plot the correlations between the original and the augmented images (call it \textit{correlation curve}) and the top-1000 intersections (\textit{top-1000 curve}). These curves can be visualised as in \cref{fig:curves_cor}. To compare explainability methods in a fair way, we score relative to classification certainty. For a fixed range $[M, N]$, we compute a normalized area under the response curve for $x\in[M, N]$, or, more precisely, the portion of this area out of a rectangle with corners $[M, 0], [N, 0], [N, 1], [M, 1]$. Moreover, to be able to compare the scores of different curves and let the score depend only on the shape of the curve, we ensure that the point on the curve corresponding to the zero-augmented image takes a value of $1$ by shifting the response curve. \Cref{fig:Score-AOC} illustrates how the score is computed. For each curve, we get a number between 0 and 1 and higher values indicate a more stable response. Finally, since we want to compare the robustness of the model's predictions and its explanations, we divide the score for the correlation (or top-1000) curve of explanations by the score for the probability curve and denote it as \textit{S(corelation, probability)} (or \textit{S(top-1000, probability)}). If S($\cdot$, probability) is smaller than 1, it means that the predictions are more stable than the explanations, whereas values higher than 1 entails more robust explanations.
The intervals for augmentation parameters are chosen such that the probability of the target class drops on average by at least $10\%$ at one of the endpoints (in comparison to the original image).

Apart from comparing the robustness of the explainability methods, we are interested in the overall quality of explanations. One  method for evaluating the quality is pixel flipping \cite{Bach_Binder_Montavon_Klauschen_Muller_Samek_2015}. We consider only the original and correctly classified images. We flip the most relevant pixels first and replace them with black pixels. For each perturbed image, we divide its probability of the target class by the original image's probability of the target class and plot these values as a curve by linear interpolation. We compute the normalized area over the curve (up to $1$) from zero to the first $20\%$ pixels flipped and average these numbers across all images. \Cref{fig:PF-AOC} visualizes how the pixel flipping score is computed. A similar definition has been given by Samek \etal \cite{Samek_Binder_Montavon_Lapuschkin_Muller_2017}. Our definition differs in dividing the probabilities instead of subtracting them. The fractions better capture the relative decline of the probability and can take all values in $[0, 1]$.
\vspace{-3mm}
\paragraph{Networks}
We study three convolutional networks (ResNet50 \cite{he2016deep}, VGG16 \cite{simonyan2014very}, and EfficientNetV2 small \cite{tan2021efficientnetv2}). Because of space constraints, we present in this paper only the results for ResNet50. However, the results for VGG16 and EfficientNet V2 small show similar tendencies.
Since we wanted to explore the role of augmenting images during training, we trained each model architecture with two different settings. Models trained with fully augmented data (denoted "full aug" in the following) were trained with Trivial Augment wide \cite{Muller_Hutter_2021} strategy. Models trained with limited data augmentation (denoted "lim aug") used only random resized cropping, random horizontal flipping and random erasing \cite{zhong2020random} (only EfficientNet V2 and ResNet50). Details on training can be found in \cref{appendix:training}.
\paragraph{Explanation methods}
We investigated the following explanation methods: Gradients \cite{simonyan2013deep}, Input x Gradients \cite{simonyan2013deep}, Integrated Gradients \cite{sundararajan2017axiomatic},  Guided Backpropagation \cite{springenberg2014striving}, Deconvolution \cite{zeiler2014visualizing} and three variants of Layer-wise Relevance Propagation \cite{bach2015pixel, kohlbrenner2020towards} composites:
EpsilonPlusFlat (LRP-$\varepsilon$-rule for dense layers, LRP-$\alpha, \beta$ ($\alpha=1, \beta=0$), also called ZPlus rule, for convolutional layers, and the flat rule for the first linear layer), 
EpsilonGammaBox (LRP-$\varepsilon$-rule for dense layers, the LRP-$\gamma$-rule ($\gamma=0.25)$ for
convolutional layers, and the LRP-$Z^B$-rule (or box-rule) for the first layer) and 
EpsilonAlpha2Beta1Flat (LRP-$\varepsilon$-rule for dense layers, LRP-$\alpha, \beta$ ($\alpha=2, \beta=1$) for convolutional layers and the flat rule for the first linear layer) \cite{Montavon_Binder_Lapuschkin_Samek_Muller_2019}.
We used Zennit \cite{anders2021software} to generate LRP explanations and Captum \cite{kokhlikyan2020captum} for the rest of the explainability methods.

The code and hyperparameters for reproducing the experiments can be found in the project repository \footnote{\url{https://github.com/LenkaTetkova/robustness-of-explanations.git}}.
%
\section{Results}\label{sec:results}
\begin{figure}
  \centering
   \includegraphics[width=\linewidth]{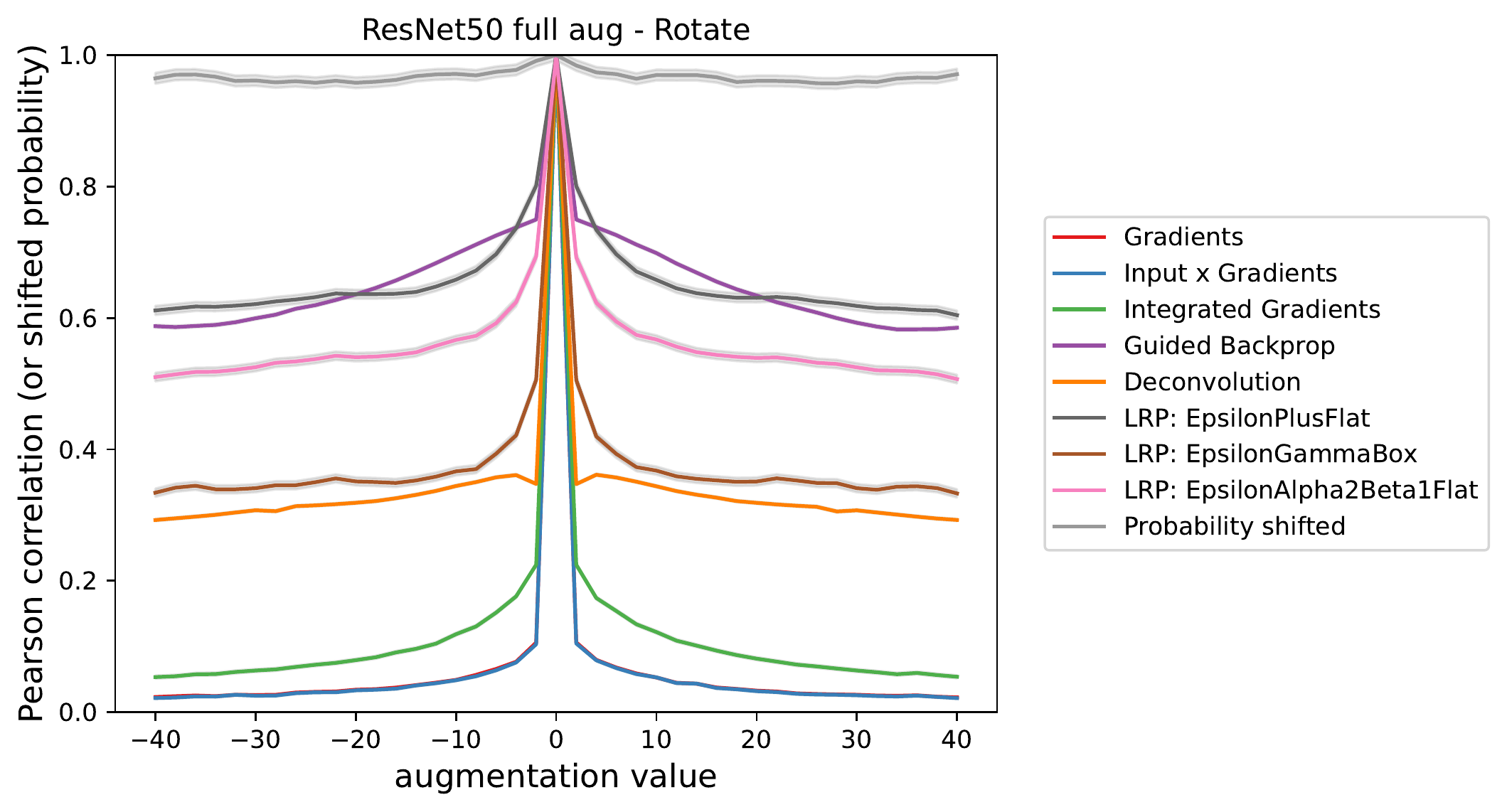}    
   \caption{Example of curves showing the probabilities and correlations between the original and the rotated images.}
   \label{fig:curves_cor}
   \vspace{-5mm}
\end{figure}
\begin{figure}
  \centering
   \includegraphics[width=\linewidth]{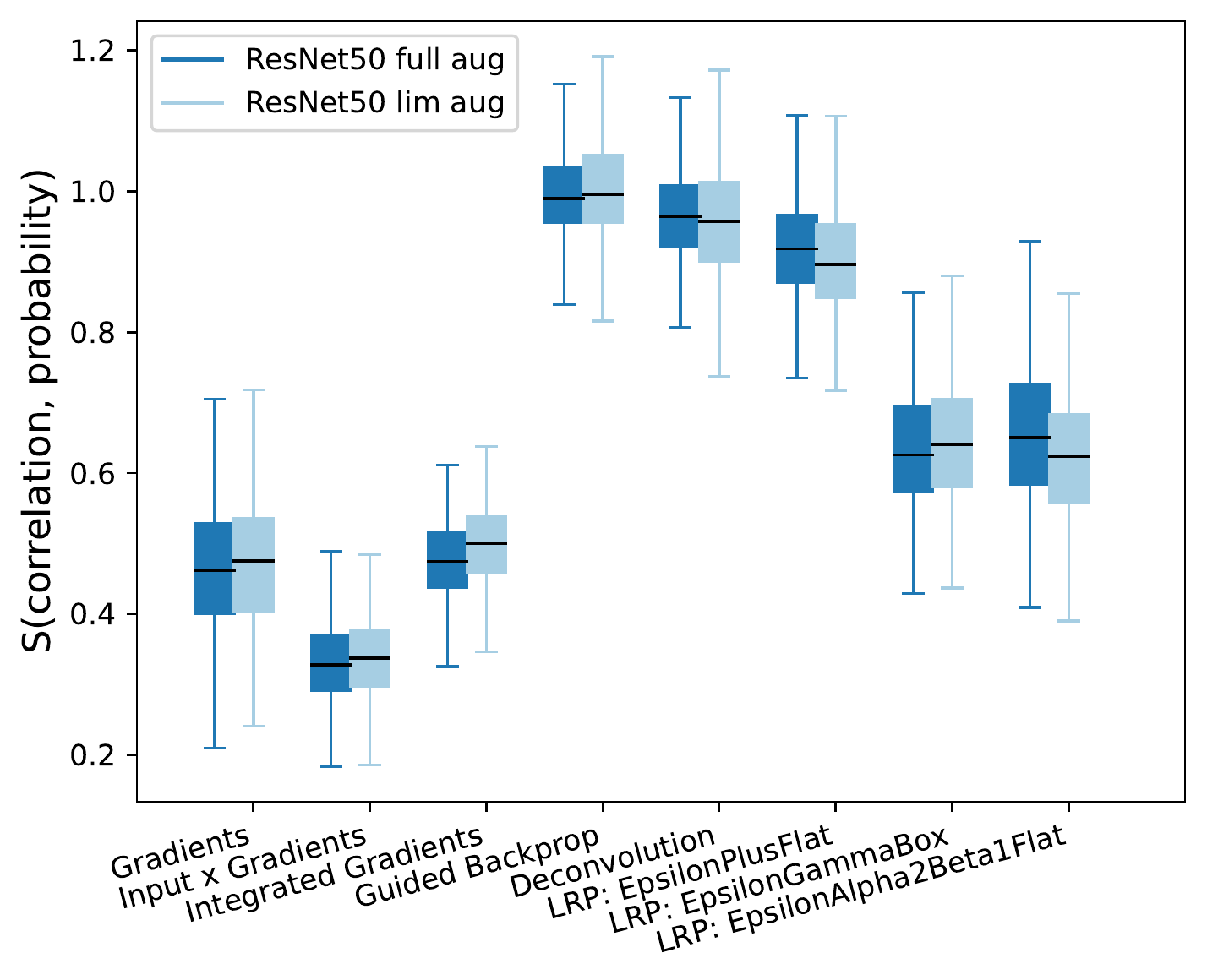}    
    \caption{Comparison of "ResNet50 full aug" (391 images) and "ResNet50 lim aug" (385 images) for each explainability method. We plot S(corelation, probability) for changes in brightness (AddToBrightness from -95 to 95). Boxes show the quartiles and medians, and whiskers extend to the most extreme, non-outlier data points.}
    \label{fig:boxplot_ResNet50_AddToBrightness_cor}
   \vspace{-5mm}
\end{figure}

\begin{table*}
  \centering
  \begin{tabular}{|l||l|l|l||l|l|l|} \hline
     & Brightness & Hue & Saturation & Rotate & Scale & Translate \\ \hline \hline
    Gradients & 0.468 & 0.442 & 0.354 & 0.127 & 0.122 & 0.246 \\ \hline 
    Input x Gradients & 0.330 & 0.443 & 0.343 & 0.126 & 0.120 & 0.245 \\ \hline 
    Integrated Gradients & 0.478 & 0.636 & 0.546 & 0.209 & 0.229 & 0.327 \\ \hline
    Guided Backprop & \textbf{1.005} & 1.028 & 0.994 & \textbf{0.819} & \textbf{0.866} & \textbf{0.875} \\ \hline 
    Deconvolution & 0.975 & 1.014 & 0.975 & 0.434 & 0.437 & 0.449 \\ \hline
    LRP: EpsilonPlusFlat & 0.923 & \textbf{1.053} & \textbf{1.038} & 0.796 & 0.834 & 0.792 \\ \hline
    LRP: EpsilonGammaBox & 0.632 & 0.856 & 0.832 & 0.480 & 0.512 & 0.532 \\ \hline
    LRP: EpsilonAlpha2Beta1Flat & 0.662 & 1.006 & 0.972 & 0.691 & 0.722 & 0.706 \\ \hline
  \end{tabular}
  \caption{Results of S(correlation, probability) for "ResNet50 full aug", computed on 391 (correctly classified) images. All numbers are with uncertainty (standard error of the mean) at most $\pm 0.007$. Highlighted are the highest values for each augmentation.}
  \label{tab:resnet50_all}
  \vspace{-5mm}
\end{table*}
\Cref{fig:curves_cor} shows the probability and correlation curves for rotation and "ResNet50 full aug". It shows that, although the predictions do not change much for increasing magnitudes of augmentation, the drop in correlation is huge.
\Cref{tab:resnet50_all} shows S(corelation, probability) for all augmentation and explainability methods tested on "ResNet50 full aug". We observe that the explanations are in most cases less stable than the predictions. Moreover, the robustness of explanations depends on the augmentation method – for some of them, the explanations are more robust than for others. Specifically,  explanations of images augmented by invariant methods are more stable than the ones augmented by equivariant methods.
The variance in robustness across explainability methods was an unexpected finding. The most stable ones, the composites of LRP and Guided Backprop, indicate a certain degree of stability, whereas the least stable ones, Gradients and Gradients x Inputs, show a steep decrease in the similarity of explanations even for small perturbations.
\Cref{fig:boxplot_ResNet50_AddToBrightness_cor} depicts the comparison of ResNet50 trained with full and limited augmentations evaluated on the changes in brightness. We can observe negligible differences between both networks. Therefore, it indicates that training with data augmentations does not diminish this problem. Additional plots for other augmentation methods can be found in \cref{appendix:all_vs_no}. 
\begin{figure}
  \centering
   \includegraphics[width=\linewidth]{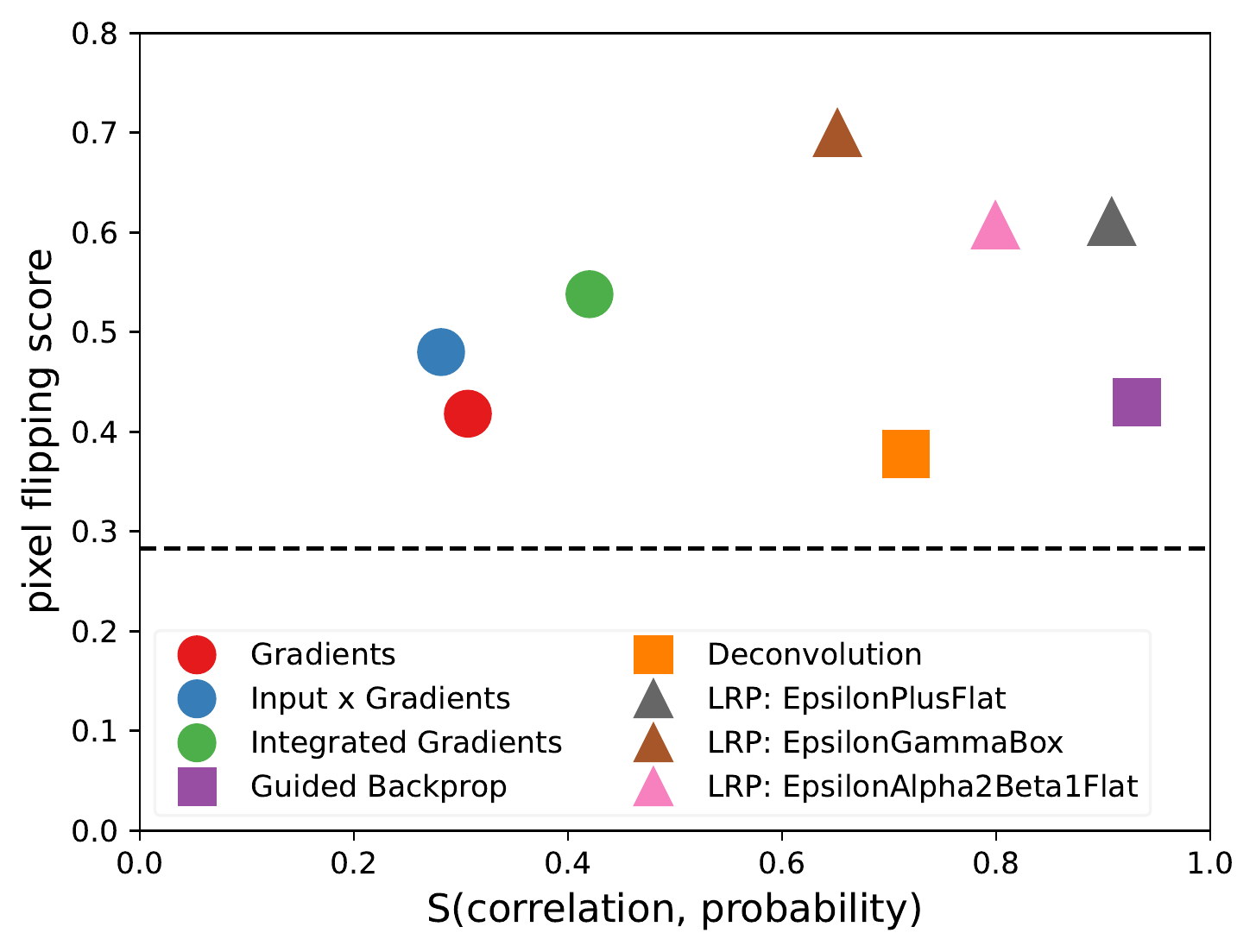}    
    \caption{Comparison of S(corelation, probability) and pixel flipping score for "ResNet50 full aug". The scores are defined in \cref{sec:methods_metrics}. The x-axis shows the average of the S(corelation, probability) for all six augmentation methods used in this paper. The dashed line corresponds to a baseline pixel-flipping score computed with random sorting of the pixels. The best methods are in the top right corner.}
   \label{fig:PF_ResNet50_all}
   \vspace{-5mm}
\end{figure}
However, stability is not the only desired property of explainability methods. We need to consider also their overall quality. In our experiments, we measured faithfulness, specifically pixel flipping score. \Cref{fig:PF_ResNet50_all} shows S(corelation, probability) against the pixel-flipping scores. We can observe that all the LRP composites lie in the top-right corner. On the other hand, Guided Backprop and Deconvolution attain low pixel-flipping scores in comparison to other methods. This is not surprising because Nie \etal \cite{Nie_Zhang_Patel_2020} showed that these two methods do not depend much on the tested model but rather perform a (partial) image recovery.

We consider the instability of explanations to be a serious problem that is relevant for many domains where computer vision tasks are solved using neural networks. Our study contributes additional evidence that current explainability methods cannot be trusted to deliver a reliable justification of the outputs of a model. Many of the tested perturbations may occur unintentionally when taking images under different light conditions, from a different angle or by domain shift and variability of the data. Unless more stability of explainability methods is ensured, explanations cannot be trusted and used as a foundation for authorizing neural networks with important tasks with significant impact.
%
\section{Conclusion}
We investigated the robustness of post-hoc explainability methods under natural perturbations of the input images. We found out that LRP composites and Guided Backprop produce the most stable explanations and Gradients and Input x Gradients are the least stable ones. When perturbing with the invariant methods (\eg, changing brightness, hue and saturation), the explanations are more stable than when perturbing with equivariant methods (\eg, rotation, scaling and translation). Training with data augmentation does not reduce this problem.
%
\section{Acknowledgements}
This work was supported by the DIREC Bridge project Deep Learning and Automation of Imaging‐Based Quality of Seeds and Grains, Innovation Fund Denmark grant number 9142‐00001B and by the Danish Pioneer Centre for AI, DNRF grant number P1. We acknowledge EuroHPC Joint Undertaking for awarding us access to Karolina at IT4Innovations, Czech Republic.
{\small
\bibliographystyle{ieeetr}
\bibliography{arxiv}

\begin{thebibliography}{10}

\bibitem{goodfellow2014explaining}
I.~J. Goodfellow, J.~Shlens, and C.~Szegedy, ``Explaining and harnessing
  adversarial examples,'' {\em arXiv preprint arXiv:1412.6572}, 2014.

\bibitem{dombrowski2019explanations}
A.-K. Dombrowski, M.~Alber, C.~Anders, M.~Ackermann, K.-R. M{\"u}ller, and
  P.~Kessel, ``Explanations can be manipulated and geometry is to blame,'' {\em
  Advances in Neural Information Processing Systems}, vol.~32, 2019.

\bibitem{ghorbani2019interpretation}
A.~Ghorbani, A.~Abid, and J.~Zou, ``Interpretation of neural networks is
  fragile,'' in {\em Proceedings of the AAAI conference on artificial
  intelligence}, vol.~33, pp.~3681--3688, 2019.

\bibitem{akhtar2018threat}
N.~Akhtar and A.~Mian, ``Threat of adversarial attacks on deep learning in
  computer vision: A survey,'' {\em Ieee Access}, vol.~6, pp.~14410--14430,
  2018.

\bibitem{NIPS1996_81e5f81d}
L.~Yaeger, R.~Lyon, and B.~Webb, ``Effective training of a neural network
  character classifier for word recognition,'' in {\em Advances in Neural
  Information Processing Systems} (M.~Mozer, M.~Jordan, and T.~Petsche, eds.),
  vol.~9, MIT Press, 1996.

\bibitem{Cire_an_2010}
D.~C. Cire{\c{s}}an, U.~Meier, L.~M. Gambardella, and J.~Schmidhuber, ``Deep,
  big, simple neural nets for handwritten digit recognition,'' {\em Neural
  Computation}, vol.~22, pp.~3207--3220, dec 2010.

\bibitem{wang2017effectiveness}
J.~Wang, L.~Perez, {\em et~al.}, ``The effectiveness of data augmentation in
  image classification using deep learning,'' {\em Convolutional Neural
  Networks Vis. Recognit}, vol.~11, no.~2017, pp.~1--8, 2017.

\bibitem{shorten2019survey}
C.~Shorten and T.~M. Khoshgoftaar, ``A survey on image data augmentation for
  deep learning,'' {\em Journal of big data}, vol.~6, no.~1, pp.~1--48, 2019.

\bibitem{rebuffi2021data}
S.-A. Rebuffi, S.~Gowal, D.~A. Calian, F.~Stimberg, O.~Wiles, and T.~A. Mann,
  ``Data augmentation can improve robustness,'' {\em Advances in Neural
  Information Processing Systems}, vol.~34, pp.~29935--29948, 2021.

\bibitem{Won_Bae_Kim_2023}
S.~Won, S.-H. Bae, and S.~T. Kim, ``Analyzing effects of mixed sample data
  augmentation on model interpretability,'' Mar 2023.
\newblock arXiv:2303.14608 [cs].

\bibitem{wang2021selfinterpretable}
Y.~Wang and X.~Wang, ``Self-interpretable model with transformation equivariant
  interpretation,'' in {\em Advances in Neural Information Processing Systems}
  (A.~Beygelzimer, Y.~Dauphin, P.~Liang, and J.~W. Vaughan, eds.), 2021.

\bibitem{Swartout_Moore_1993}
W.~Swartout and J.~Moore, ``Explanation in second generation expert systems,''
  p.~543–585, Jan 1993.

\bibitem{Adebayo_Gilmer_Muelly_Goodfellow_Hardt_Kim_2018}
J.~Adebayo, J.~Gilmer, M.~Muelly, I.~Goodfellow, M.~Hardt, and B.~Kim, ``Sanity
  checks for saliency maps,'' in {\em Advances in Neural Information Processing
  Systems}, vol.~31, Curran Associates, Inc., 2018.

\bibitem{Kindermans_Hooker_Adebayo_Alber_Schutt_Dahne_Erhan_Kim_2017}
P.-J. Kindermans, S.~Hooker, J.~Adebayo, M.~Alber, K.~T. Schütt, S.~Dähne,
  D.~Erhan, and B.~Kim, ``The (un)reliability of saliency methods,'' Nov 2017.
\newblock arXiv:1711.00867 [cs, stat].

\bibitem{Nie_Zhang_Patel_2020}
W.~Nie, Y.~Zhang, and A.~Patel, ``A theoretical explanation for perplexing
  behaviors of backpropagation-based visualizations,'' Feb 2020.
\newblock arXiv:1805.07039 [cs].

\bibitem{Bach_Binder_Montavon_Klauschen_Muller_Samek_2015}
S.~Bach, A.~Binder, G.~Montavon, F.~Klauschen, K.-R. Müller, and W.~Samek,
  ``On pixel-wise explanations for non-linear classifier decisions by
  layer-wise relevance propagation,'' {\em PLOS ONE}, vol.~10, no.~7,
  p.~e0130140, 2015.

\bibitem{Rieger_Hansen_2020}
L.~Rieger and L.~K. Hansen, ``Irof: a low resource evaluation metric for
  explanation methods,'' Mar 2020.
\newblock arXiv:2003.08747 [cs].

\bibitem{Chalasani_Chen_Chowdhury_Jha_Wu_2020}
P.~Chalasani, J.~Chen, A.~R. Chowdhury, S.~Jha, and X.~Wu, ``Concise
  explanations of neural networks using adversarial training,'' Jul 2020.
\newblock arXiv:1810.06583 [cs, stat].

\bibitem{hedstrom2023quantus}
A.~Hedstr{\"{o}}m, L.~Weber, D.~Krakowczyk, D.~Bareeva, F.~Motzkus, W.~Samek,
  S.~Lapuschkin, and M.~M.~M. H{\"{o}}hne, ``Quantus: An explainable ai toolkit
  for responsible evaluation of neural network explanations and beyond,'' {\em
  Journal of Machine Learning Research}, vol.~24, no.~34, pp.~1--11, 2023.

\bibitem{imgaug}
A.~B. Jung, K.~Wada, J.~Crall, S.~Tanaka, J.~Graving, C.~Reinders, S.~Yadav,
  J.~Banerjee, G.~Vecsei, A.~Kraft, Z.~Rui, J.~Borovec, C.~Vallentin,
  S.~Zhydenko, K.~Pfeiffer, B.~Cook, I.~Fernández, F.-M. De~Rainville, C.-H.
  Weng, A.~Ayala-Acevedo, R.~Meudec, M.~Laporte, {\em et~al.}, ``{imgaug}.''
  \url{https://github.com/aleju/imgaug}, 2020.
\newblock Online; accessed 01-Feb-2020.

\bibitem{imagenet_cvpr09}
J.~Deng, W.~Dong, R.~Socher, L.-J. Li, K.~Li, and L.~Fei-Fei, ``{ImageNet: A
  Large-Scale Hierarchical Image Database},'' in {\em CVPR09}, 2009.

\bibitem{Samek_Binder_Montavon_Lapuschkin_Muller_2017}
W.~Samek, A.~Binder, G.~Montavon, S.~Lapuschkin, and K.-R. Muller, ``Evaluating
  the visualization of what a deep neural network has learned,'' {\em IEEE
  Transactions on Neural Networks and Learning Systems}, vol.~28,
  p.~2660–2673, Nov 2017.

\bibitem{he2016deep}
K.~He, X.~Zhang, S.~Ren, and J.~Sun, ``Deep residual learning for image
  recognition,'' in {\em Proceedings of the IEEE conference on computer vision
  and pattern recognition}, pp.~770--778, 2016.

\bibitem{simonyan2014very}
K.~Simonyan and A.~Zisserman, ``Very deep convolutional networks for
  large-scale image recognition,'' {\em arXiv preprint arXiv:1409.1556}, 2014.

\bibitem{tan2021efficientnetv2}
M.~Tan and Q.~Le, ``Efficientnetv2: Smaller models and faster training,'' in
  {\em International Conference on Machine Learning}, pp.~10096--10106, PMLR,
  2021.

\bibitem{Muller_Hutter_2021}
S.~G. Müller and F.~Hutter, ``Trivialaugment: Tuning-free yet state-of-the-art
  data augmentation,'' p.~754–762, IEEE Computer Society, Oct 2021.

\bibitem{zhong2020random}
Z.~Zhong, L.~Zheng, G.~Kang, S.~Li, and Y.~Yang, ``Random erasing data
  augmentation,'' in {\em Proceedings of the AAAI conference on artificial
  intelligence}, vol.~34, pp.~13001--13008, 2020.

\bibitem{simonyan2013deep}
K.~Simonyan, A.~Vedaldi, and A.~Zisserman, ``Deep inside convolutional
  networks: Visualising image classification models and saliency maps,'' {\em
  arXiv preprint arXiv:1312.6034}, 2013.

\bibitem{sundararajan2017axiomatic}
M.~Sundararajan, A.~Taly, and Q.~Yan, ``Axiomatic attribution for deep
  networks,'' in {\em International conference on machine learning},
  pp.~3319--3328, PMLR, 2017.

\bibitem{springenberg2014striving}
J.~T. Springenberg, A.~Dosovitskiy, T.~Brox, and M.~Riedmiller, ``Striving for
  simplicity: The all convolutional net,'' {\em arXiv preprint
  arXiv:1412.6806}, 2014.

\bibitem{zeiler2014visualizing}
M.~D. Zeiler and R.~Fergus, ``Visualizing and understanding convolutional
  networks,'' in {\em European conference on computer vision}, pp.~818--833,
  Springer, 2014.

\bibitem{bach2015pixel}
S.~Bach, A.~Binder, G.~Montavon, F.~Klauschen, K.-R. M{\"u}ller, and W.~Samek,
  ``On pixel-wise explanations for non-linear classifier decisions by
  layer-wise relevance propagation,'' {\em PloS one}, vol.~10, no.~7,
  p.~e0130140, 2015.

\bibitem{kohlbrenner2020towards}
M.~Kohlbrenner, A.~Bauer, S.~Nakajima, A.~Binder, W.~Samek, and S.~Lapuschkin,
  ``Towards best practice in explaining neural network decisions with lrp,'' in
  {\em 2020 International Joint Conference on Neural Networks (IJCNN)},
  pp.~1--7, IEEE, 2020.

\bibitem{Montavon_Binder_Lapuschkin_Samek_Muller_2019}
G.~Montavon, A.~Binder, S.~Lapuschkin, W.~Samek, and K.-R. Müller, {\em
  Layer-Wise Relevance Propagation: An Overview}, p.~193–209.
\newblock Sep 2019.
\newblock journalAbbreviation: Lecture Notes in Computer Science (including
  subseries Lecture Notes in Artificial Intelligence and Lecture Notes in
  Bioinformatics).

\bibitem{anders2021software}
C.~J. Anders, D.~Neumann, W.~Samek, K.-R. Müller, and S.~Lapuschkin,
  ``Software for dataset-wide xai: From local explanations to global insights
  with {Zennit}, {CoRelAy}, and {ViRelAy},'' {\em CoRR}, vol.~abs/2106.13200,
  2021.

\bibitem{kokhlikyan2020captum}
N.~Kokhlikyan, V.~Miglani, M.~Martin, E.~Wang, B.~Alsallakh, J.~Reynolds,
  A.~Melnikov, N.~Kliushkina, C.~Araya, S.~Yan, and O.~Reblitz-Richardson,
  ``Captum: A unified and generic model interpretability library for pytorch,''
  2020.

\end{thebibliography}
}

\clearpage
\appendix
\addcontentsline{toc}{section}{Appendix}
\section*{Appendix}
\section{Details of training}\label{appendix:training}
For training, we divided the original ImageNet training set into training and validation set and used the original validation set for testing, because the labels of the original test set are not available.

We used the reference scripts published by Torchvision\footnote{\url{https://github.com/pytorch/vision/tree/main/references/classification}} for training. We used the specified recipes for training the models and changed only the parameter regarding augmenting images to get the models trained with different strategies. We decided to keep these basic augmentations to obtain models with similar performance. Resized cropping could be interpreted as translating and scaling the image. However, our results do not indicate a difference between these two methods and the rest, see \cref{sec:results}.

The "ResNet50 full aug" model was trained with the following command:
\begin{lstlisting}[language=TeX]
torchrun --nproc_per_node=8 train.py \
--model resnet50 \
--batch-size 128 \
--lr 0.5 \
--lr-scheduler cosineannealinglr \
--lr-warmup-epochs 5 \
--lr-warmup-method linear \
--auto-augment ta_wide \
--epochs 600 \
--random-erase 0.1 \
--weight-decay 0.00002 \
--norm-weight-decay 0.0 \
--label-smoothing 0.1 \
--mixup-alpha 0.2 \
--cutmix-alpha 1.0 \
--train-crop-size 176 \
--model-ema \
--val-resize-size 232 \
--ra-sampler \
--ra-reps=4
\end{lstlisting}

The training of "ResNet50 lim aug" is the same but without \lstinline[language=TeX]|--auto-augment ta_wide|. \Cref{tab:models_acc} shows the number of parameters, training time and accuracies of all the models.

\begin{table*}
  \centering
  \begin{tabular}{|l|l|l|l|l|l|l|}
\hline
Model & N parameters & Time (h) & Top-1 & Reported Top-1 & Top-5 & Reported Top-5\\ \hline \hline
ResNet50 full aug & 25.6M & 39 & 80.28 & 80.86 & 95.15 & 95.43\\ \hline
ResNet50 lim aug & - & 32 & 79.89 & - & 94.97 & - \\ \hline
EfficientNetV2 small full aug & 21M & 102 & 80.99 & 84.23 & 95.16 & 96.88 \\ \hline
EfficientNetV2 small lim aug & - & 101 & 80.89 & - & 95.22 & - \\ \hline 
VGG16 bn full aug & - & 16 & 73.43 & - & 91.39 & - \\ \hline 
VGG16 bn lim aug & 138.4M & 16 & 73.49 & 73.36 & 91.61 & 91.52 \\ \hline
\end{tabular}
  \caption{Number of parameters, training time and accuracies of all the models. The models were evaluated on the ImageNet validation set. Numbers of parameters and reported accuracies were copied from \url{https://pytorch.org/vision/stable/models.html\#table-of-all-available-classification-weights}. Models marked as "lim aug" were trained only with random resized cropping, horizontal flipping and, in the case of EfficientNet V2 S and ResNet50, with random erasing \cite{zhong2020random}. Models marked as "full aug" were, on top of that, trained with data augmentation Trivial Augmentation \cite{Muller_Hutter_2021}. All models were trained on 8 GPUs.}
  \label{tab:models_acc}
\end{table*}

\section{Augmentation methods}\label{appendix:aug_methods}
\Cref{tab:aug_intervals} shows the intervals of augmentation parameters used in the evaluation of "ResNet50 full aug". \Cref{fig:aug_examples2} shows examples of the figures and explanations for the extreme points of these intervals.

\begin{table}
  \centering
  \begin{tabular}{@{}lc@{}}
    \toprule
    Name of ImgAug function & Interval \\
    \midrule
    AddToBrightness & $[-95, 95]$ \\ 
    AddToHue & $[-30, 30]$ \\ 
    AddToSaturation & $[-70, 70]$ \\ 
    Rotate & $[-18, 18]$ \\
    Scale & $[0.89, 1.11]$ \\
    Translate & $[-0.06, 0.06]$ \\
    \bottomrule
  \end{tabular}
  \caption{Augmentation methods and intervals of magnitudes of these augmentations determined by the drop in probability by $10\%$ for "ResNet50 full aug". The first three methods are invariant, the last three are equivariant.}
  \label{tab:aug_intervals}
\end{table}

\begin{figure}[htb!]
  \centering
  \caption{Examples of the augmented images and their explanations.}
   \label{fig:aug_examples2}
  \begin{subfigure}{\linewidth}
   \includegraphics[width=\linewidth]{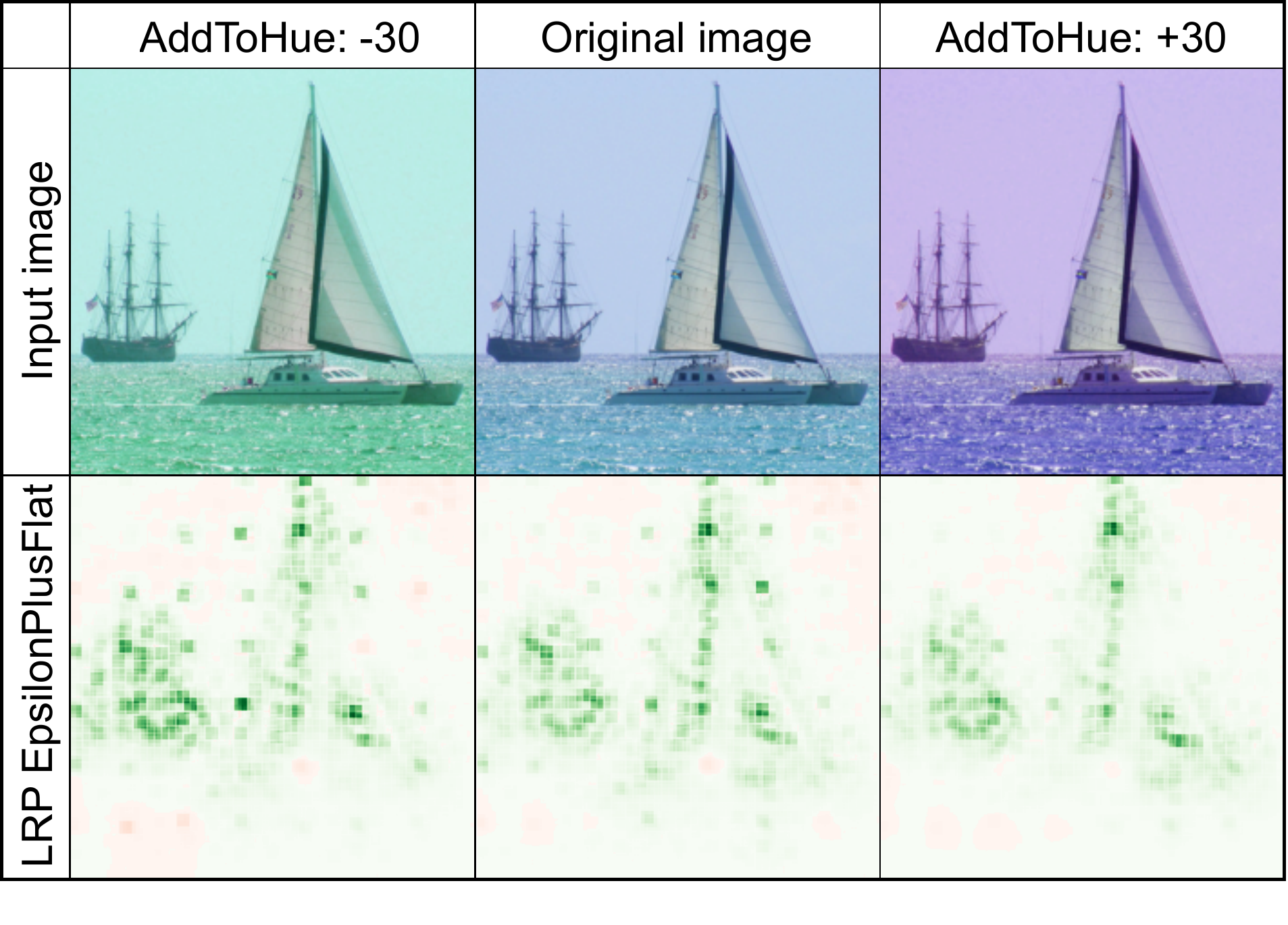}    
   \caption{Hue}
   \label{fig:aug_hue}
  \end{subfigure}
  \begin{subfigure}{\linewidth}
   \includegraphics[width=\linewidth]{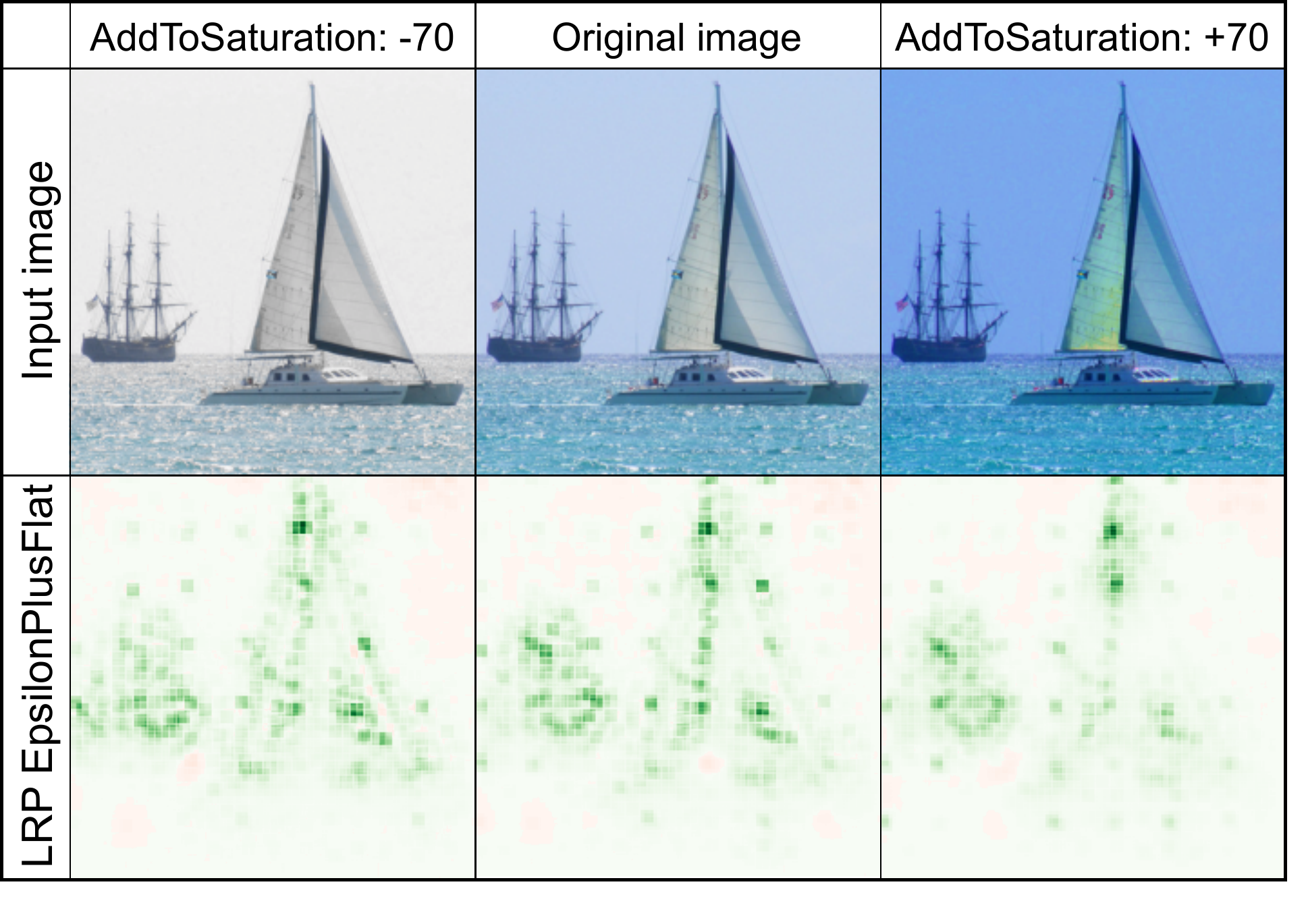}
   \caption{Saturation}
   \label{fig:aug_saturation}
  \end{subfigure}
  \begin{subfigure}{\linewidth}
   \includegraphics[width=\linewidth]{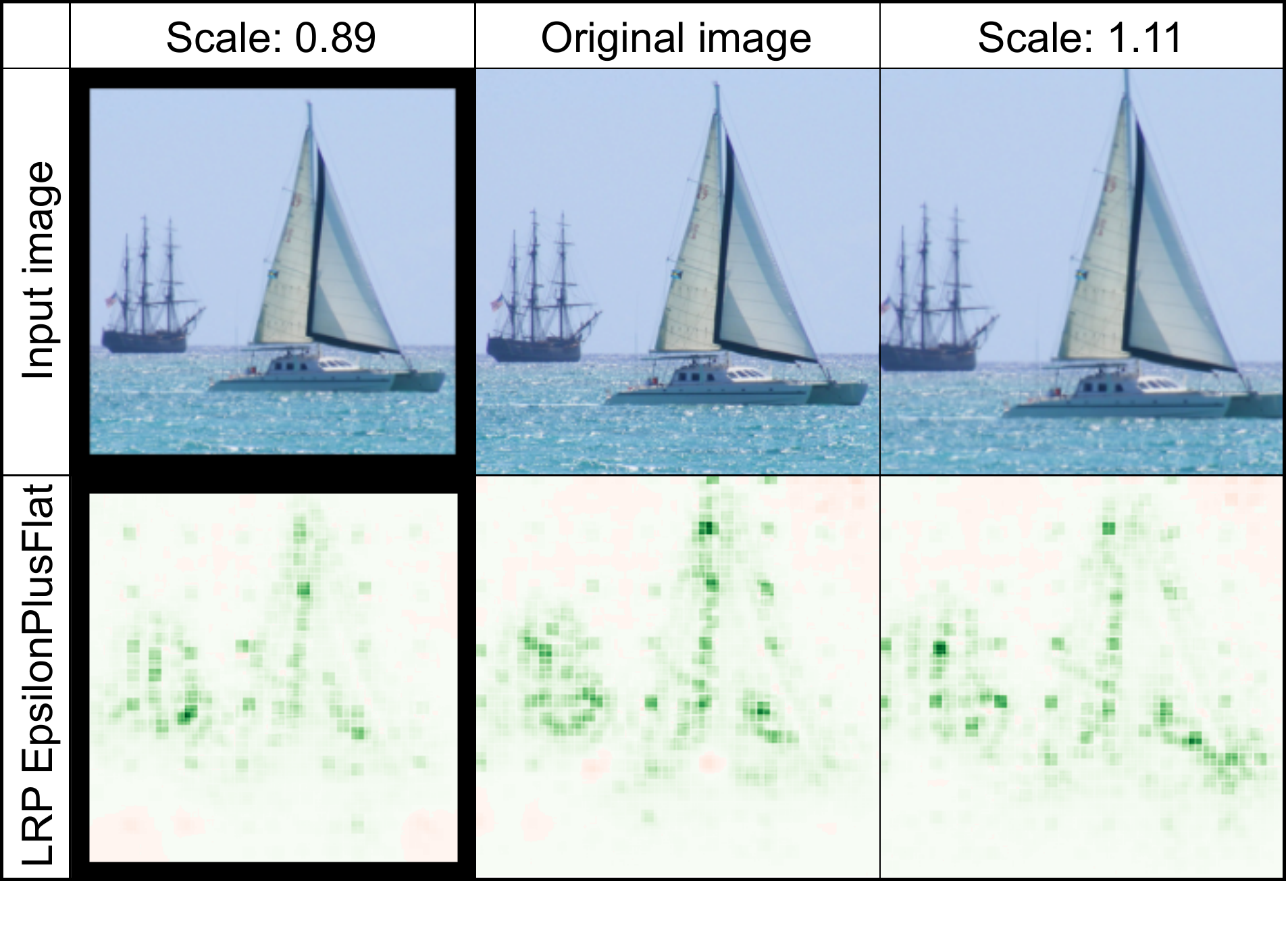}    
   \caption{Scale}
   \label{fig:aug_scale}
  \end{subfigure}
  \end{figure}
  \begin{figure}[htb!]\ContinuedFloat
  \begin{subfigure}{\linewidth}
   \includegraphics[width=\linewidth]{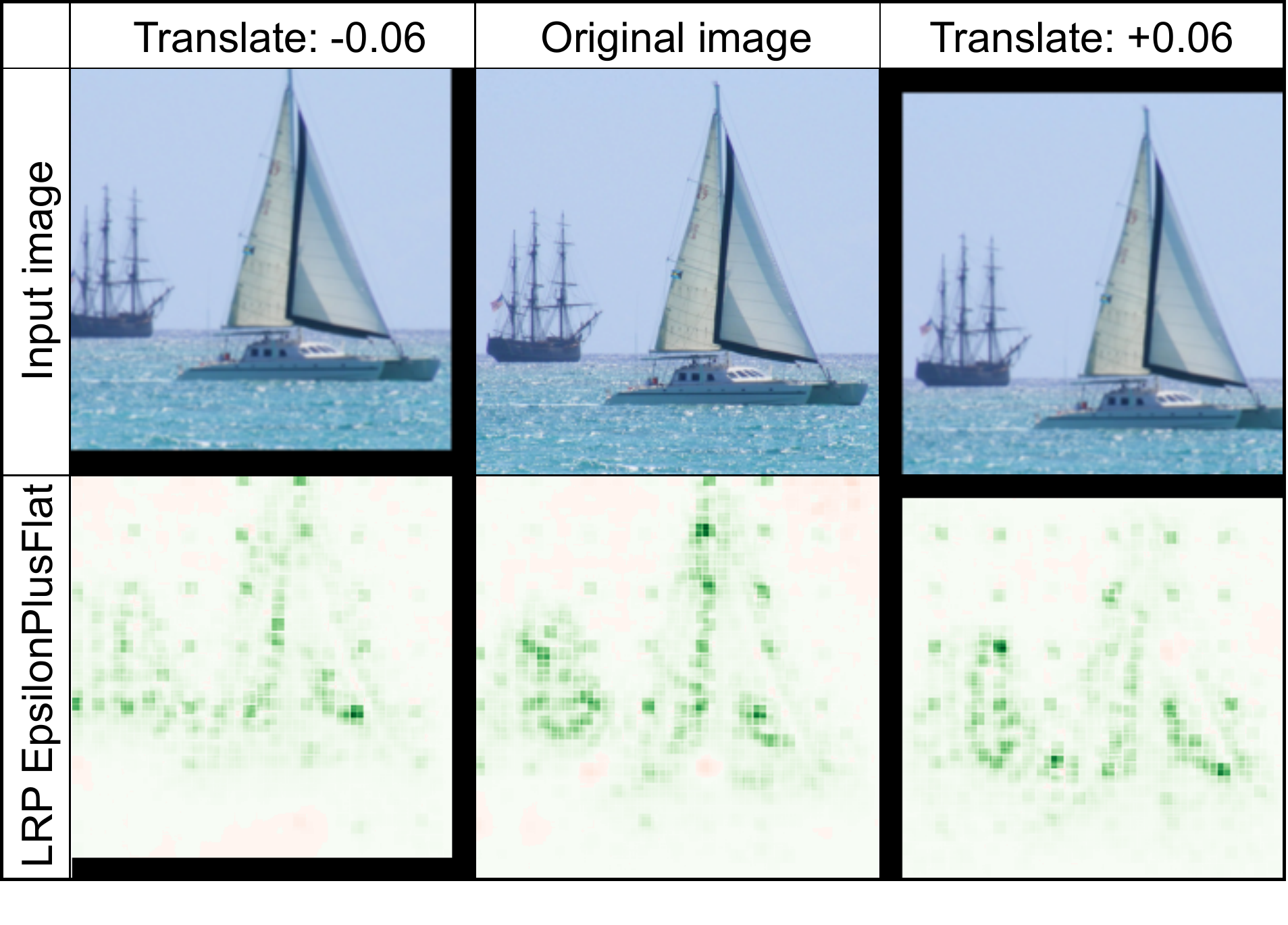}
   \caption{Translate}
   \label{fig:aug_translate}
  \end{subfigure}
\end{figure}

\section{Comparison of models trained with full and limited augmentations}\label{appendix:all_vs_no}
\Cref{fig:boxplots_ResNet50_cor} shows the comparisons in correlation and \cref{fig:boxplots_ResNet50_topk} top-1000 intersection for "ResNet50 full aug" and "ResNet50 lim aug".

\begin{figure}[!htb]
  \centering
  \caption{Comparison of ResNet50 trained with full ("full aug") and limited ("lim aug") data augmentation for each explainability method. We plot S(corelation, probability) for different perturbations. Boxes show the quartiles and medians, and whiskers extend to the most extreme, non-outlier data points.)}
  \label{fig:boxplots_ResNet50_cor}
  \begin{subfigure}{\linewidth}
   \includegraphics[width=\linewidth]{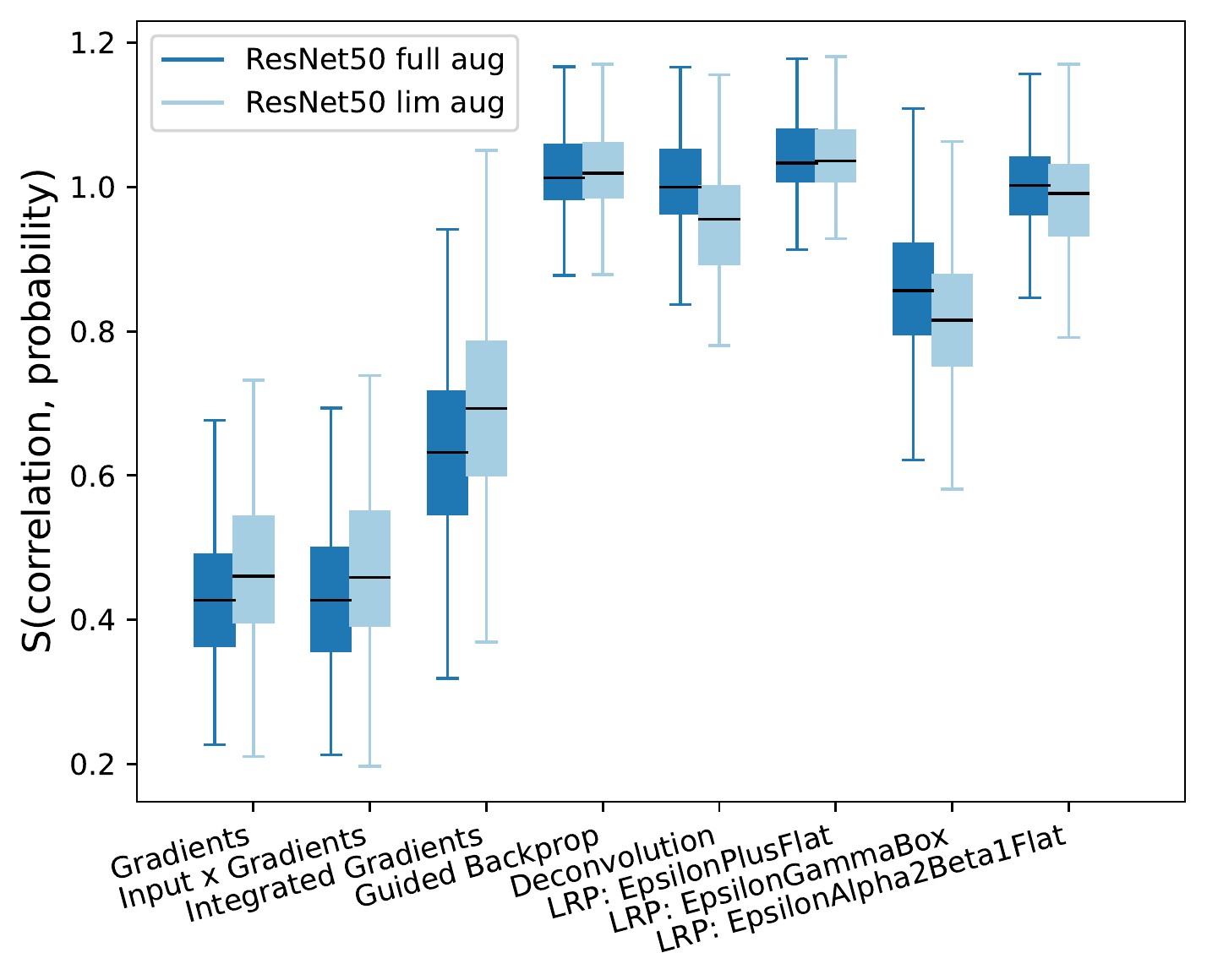}    
   \label{fig:boxplot_ResNet50_AddToHue_cor} 
   \caption{AddToHue, $[-30, 30]$}
  \end{subfigure}
  \end{figure}
  \begin{figure}\ContinuedFloat
  \begin{subfigure}{\linewidth}
      \includegraphics[width=\linewidth]{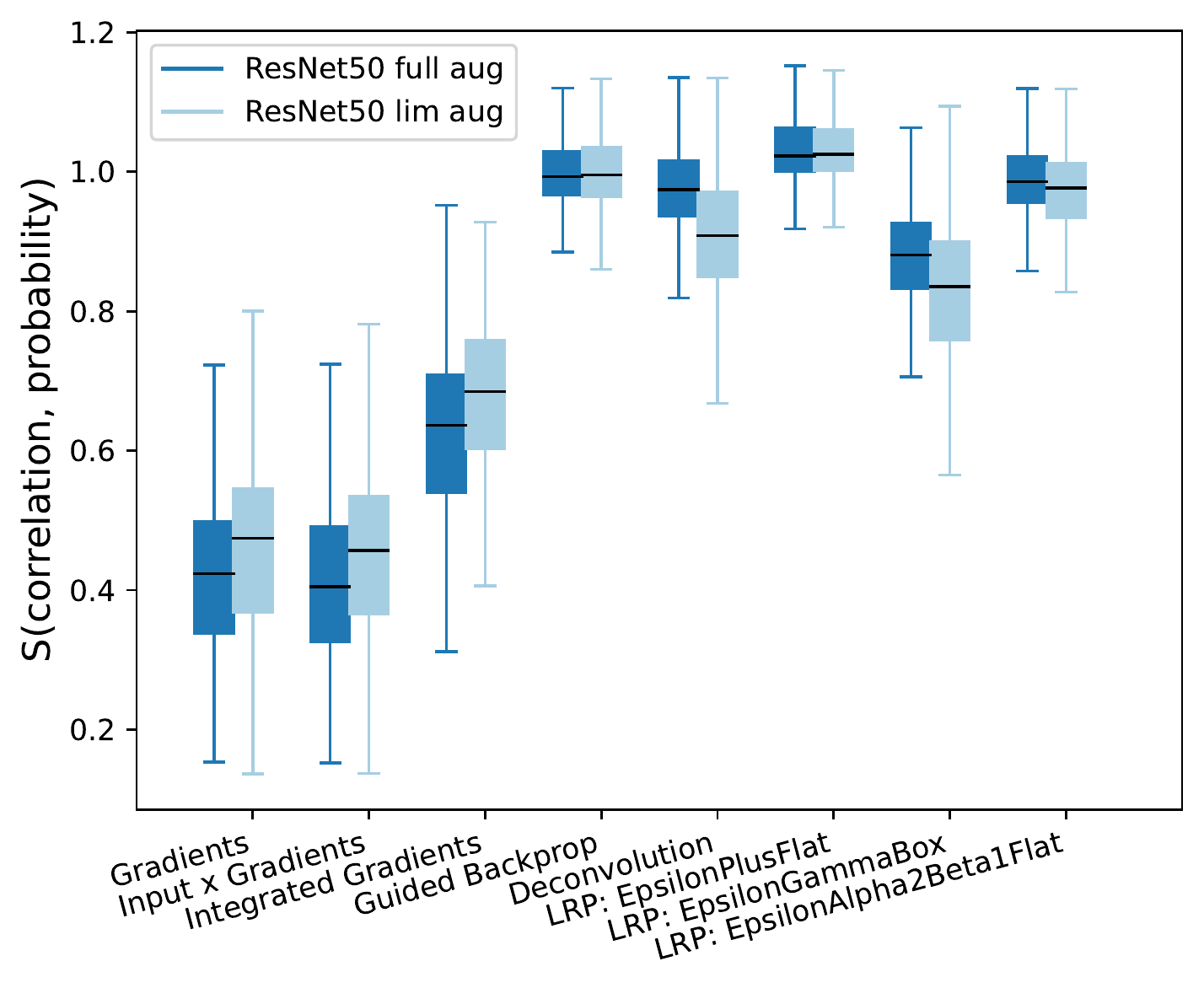}    
   \caption{AddToSaturation, $[-70, 70]$}
   \label{fig:boxplot_ResNet50_AddToSaturation_cor}
  \end{subfigure}
  \end{figure}
  \begin{figure}\ContinuedFloat
  \begin{subfigure}{\linewidth}
      \includegraphics[width=\linewidth]{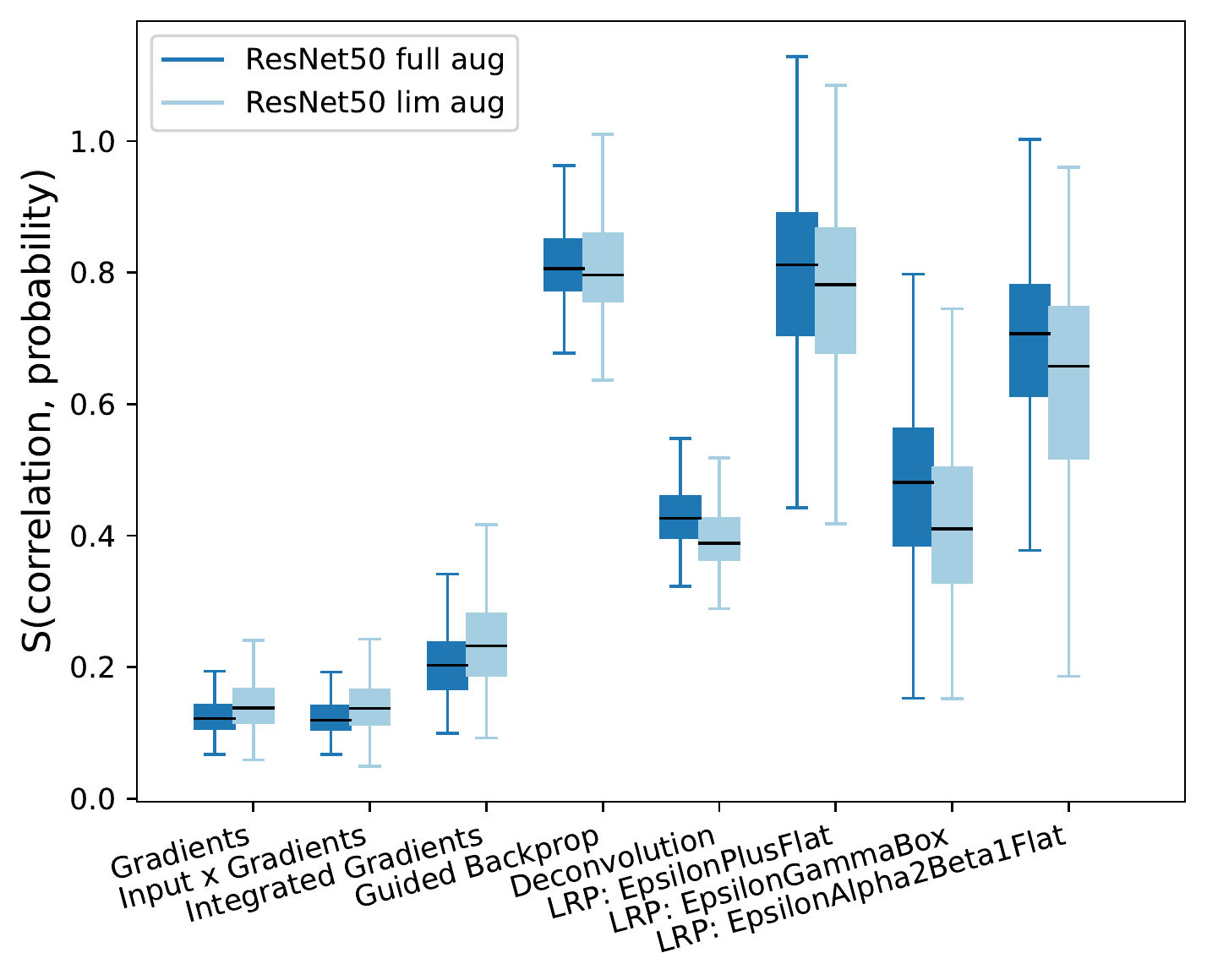}    
   \caption{Rotate, $[-18, 18]$}
   \label{fig:boxplot_ResNet50_Rotate_cor}
  \end{subfigure}
  \end{figure}
  \begin{figure}\ContinuedFloat
  \begin{subfigure}{\linewidth}
      \includegraphics[width=\linewidth]{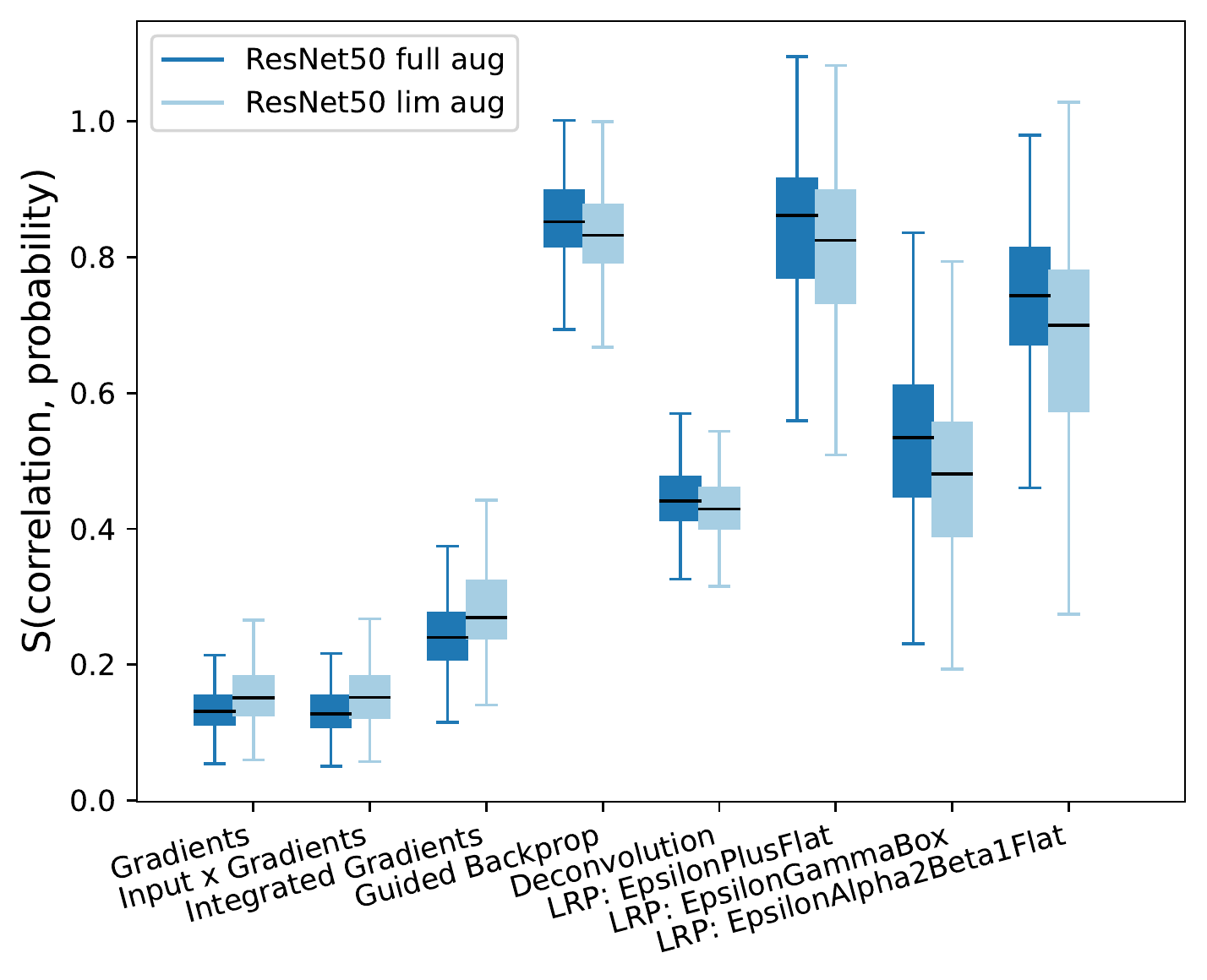}    
   \caption{Scale, $[0.89, 1.11]$}
   \label{fig:boxplot_ResNet50_Scale_cor}
  \end{subfigure}
  \end{figure}
  \begin{figure}\ContinuedFloat
  \begin{subfigure}{\linewidth}
      \includegraphics[width=\linewidth]{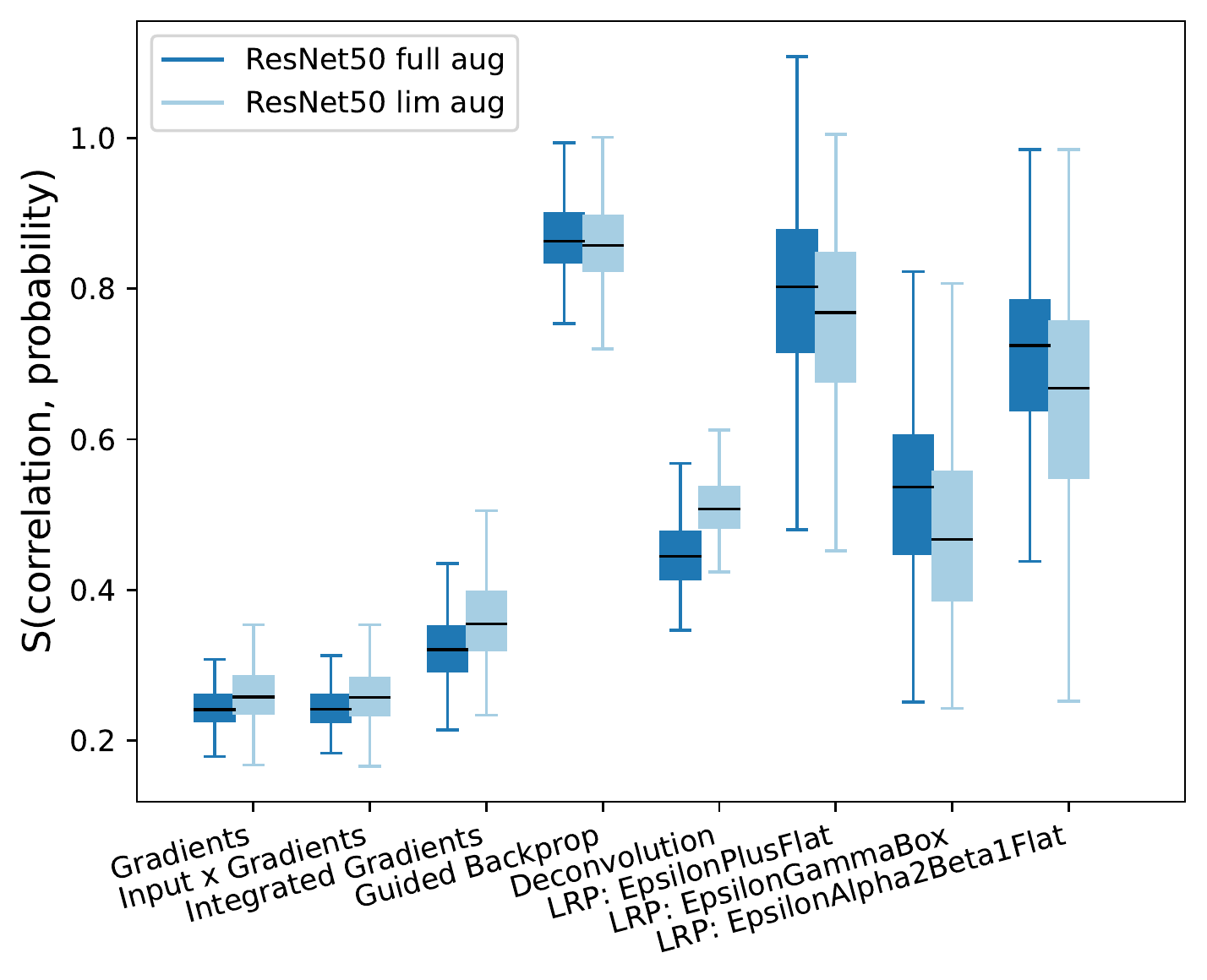}    
   \caption{Translate, $[-0.06, 0.06]$}
   \label{fig:boxplot_ResNet50_Translate_cor}
  \end{subfigure}
\end{figure}

\begin{figure}[!htb]
  \centering
  \caption{Comparison of ResNet50 trained with full ("full aug") and limiter ("lim aug") data augmentation for each explainability method. We plot S(top-1000, probability) for different perturbations. Boxes show the quartiles and medians, and whiskers extend to the most extreme, non-outlier data points.)}
  \label{fig:boxplots_ResNet50_topk}
  \begin{subfigure}{\linewidth}
   \includegraphics[width=\linewidth]{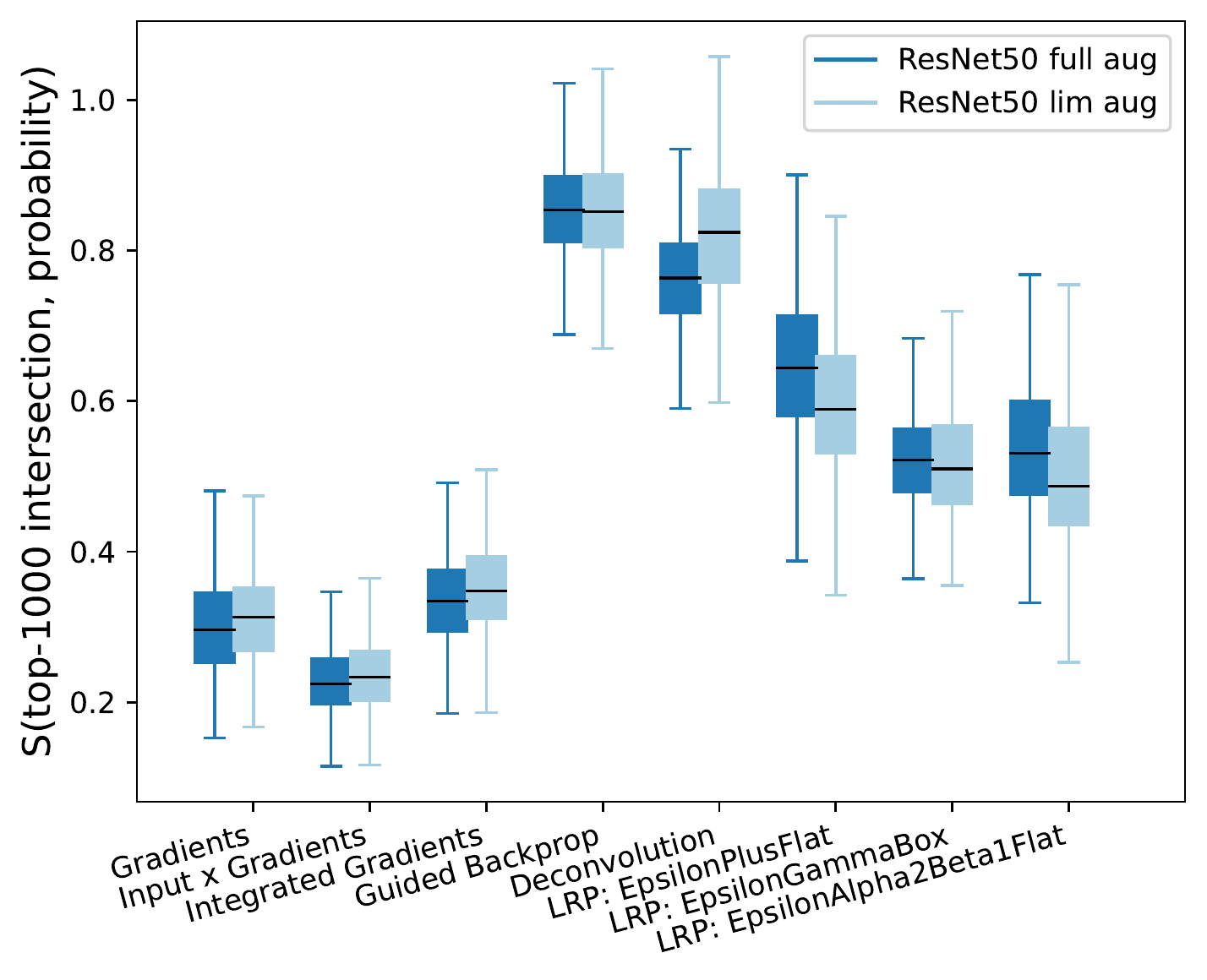}    
   \label{fig:boxplot_ResNet50_AddToBrightness_topk} 
   \caption{AddToBrightness, $[-95, 95]$}
  \end{subfigure}
  \end{figure}
  \begin{figure}\ContinuedFloat
  \begin{subfigure}{\linewidth}
   \includegraphics[width=\linewidth]{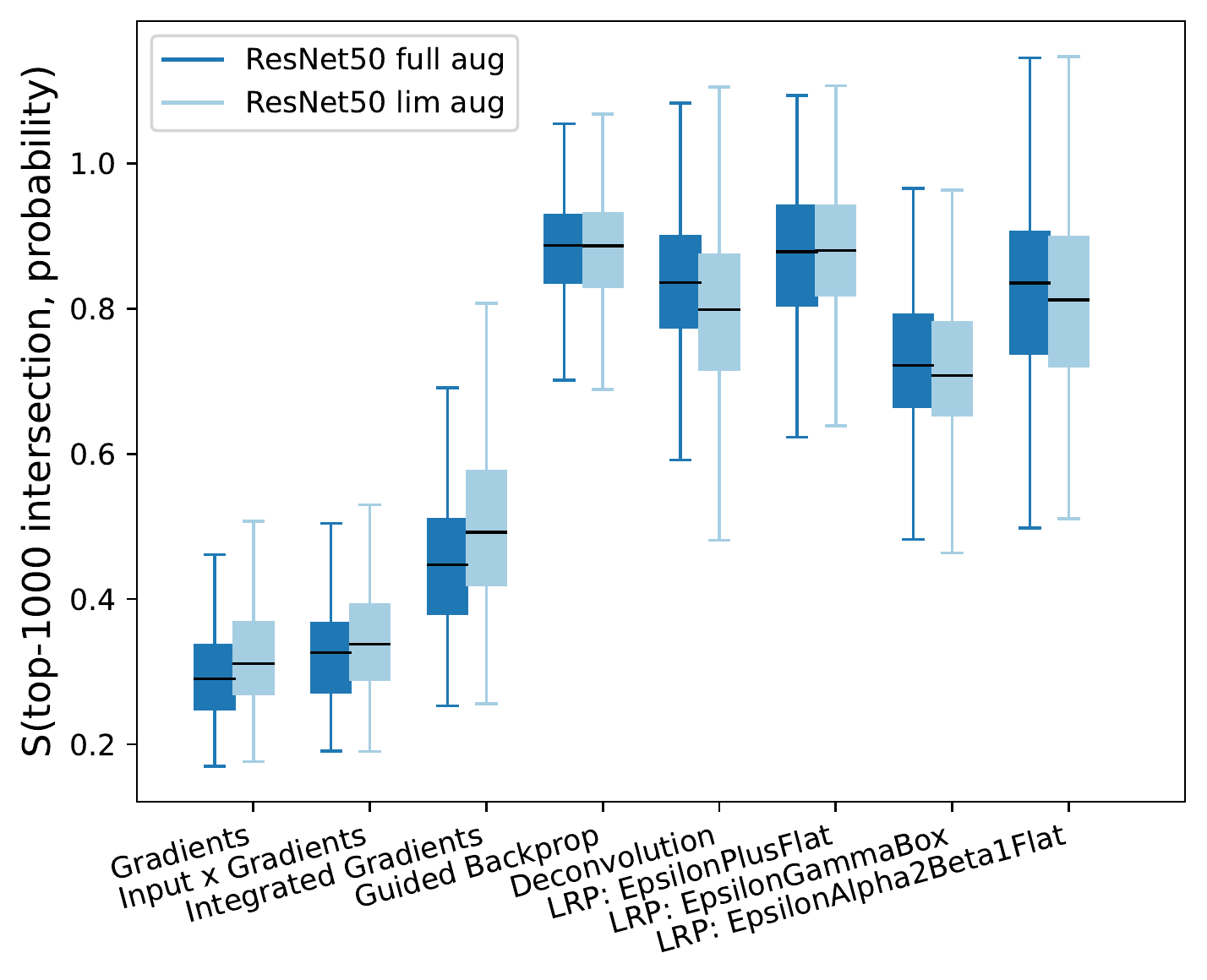}    
   \label{fig:boxplot_ResNet50_AddToHue_topk} 
   \caption{AddToHue, $[-30, 30]$}
  \end{subfigure}
  \end{figure}
  \begin{figure}\ContinuedFloat
  \begin{subfigure}{\linewidth}
      \includegraphics[width=\linewidth]{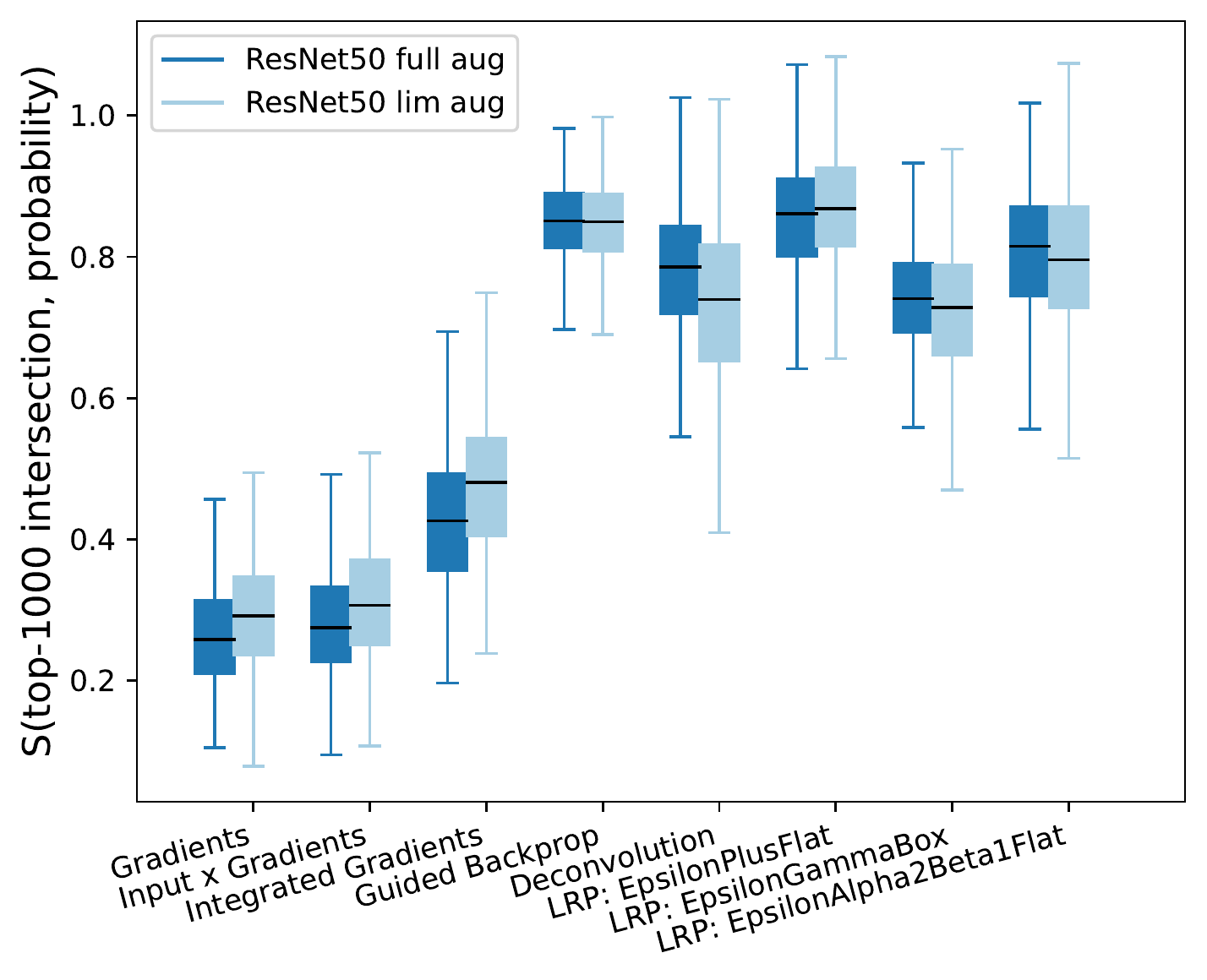}    
   \caption{AddToSaturation, $[-70, 70]$}
   \label{fig:boxplot_ResNet50_AddToSaturation_topk}
  \end{subfigure}
  \end{figure}
  \begin{figure}\ContinuedFloat
  \begin{subfigure}{\linewidth}
      \includegraphics[width=\linewidth]{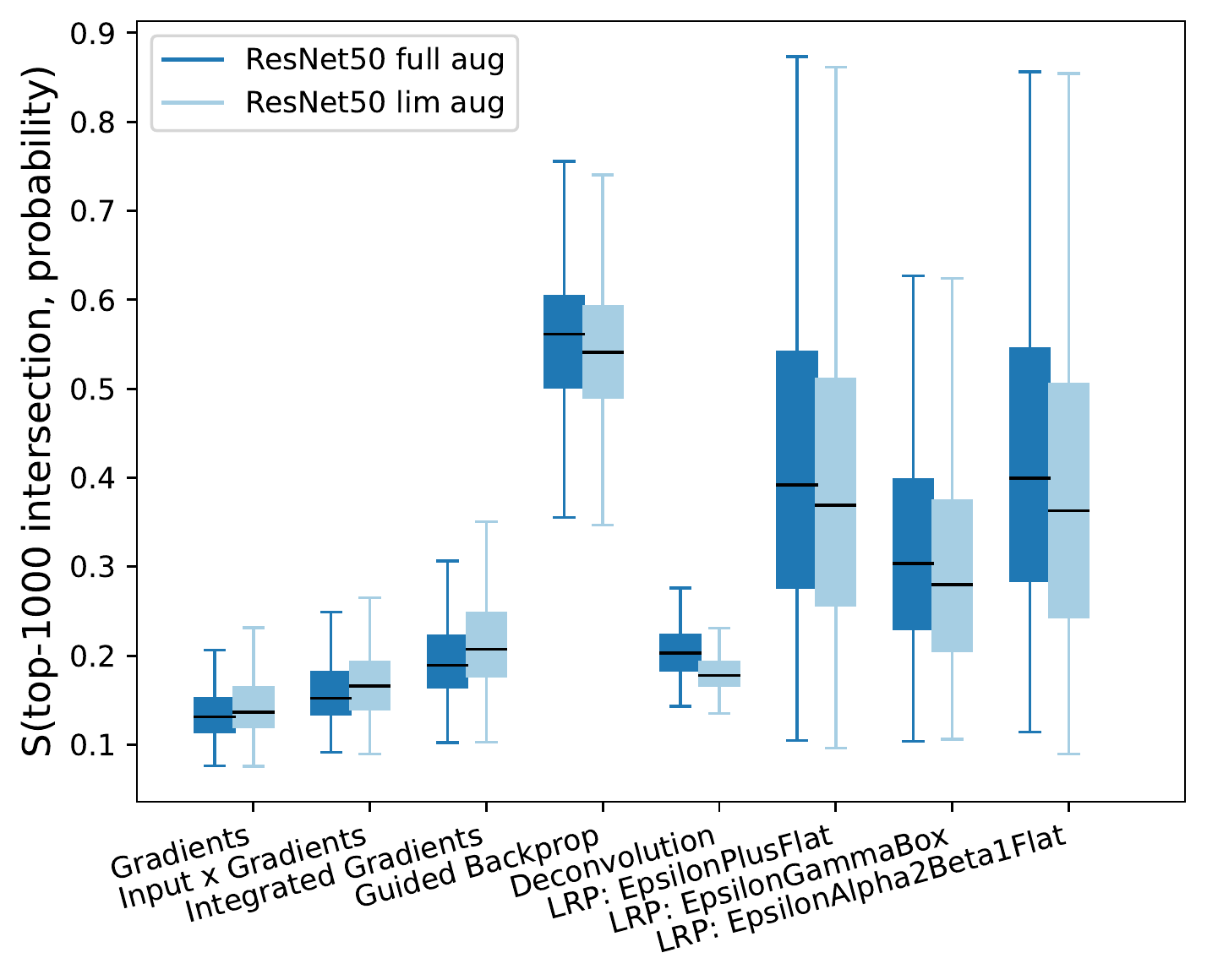}    
   \caption{Rotate, $[-18, 18]$}
   \label{fig:boxplot_ResNet50_Rotate_topk}
  \end{subfigure}
  \end{figure}
  \begin{figure}\ContinuedFloat
  \begin{subfigure}{\linewidth}
      \includegraphics[width=\linewidth]{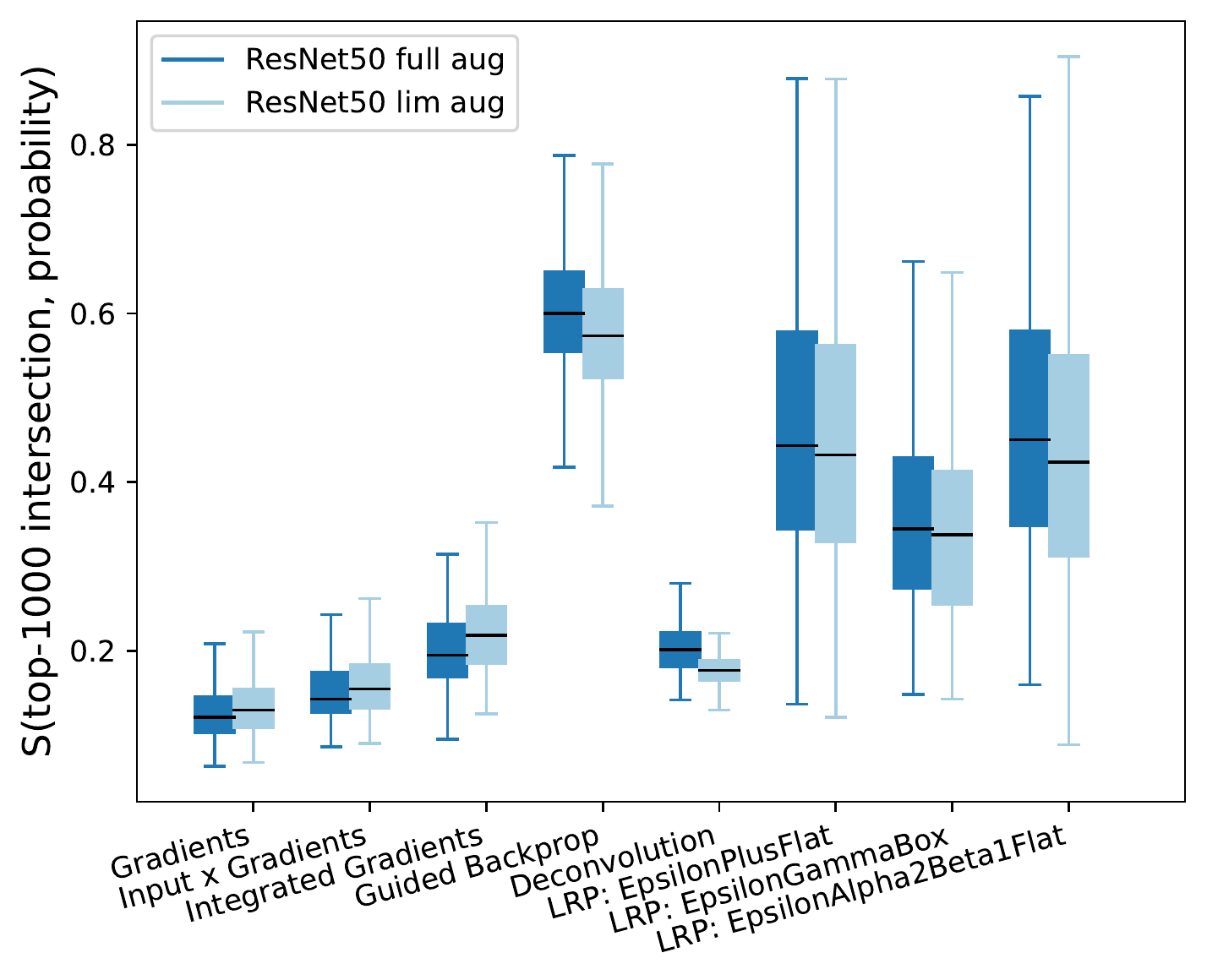}    
   \caption{Scale, $[0.89, 1.11]$}
   \label{fig:boxplot_ResNet50_Scale_topk}
  \end{subfigure}
  \end{figure}
  \begin{figure}\ContinuedFloat
  \begin{subfigure}{\linewidth}
      \includegraphics[width=\linewidth]{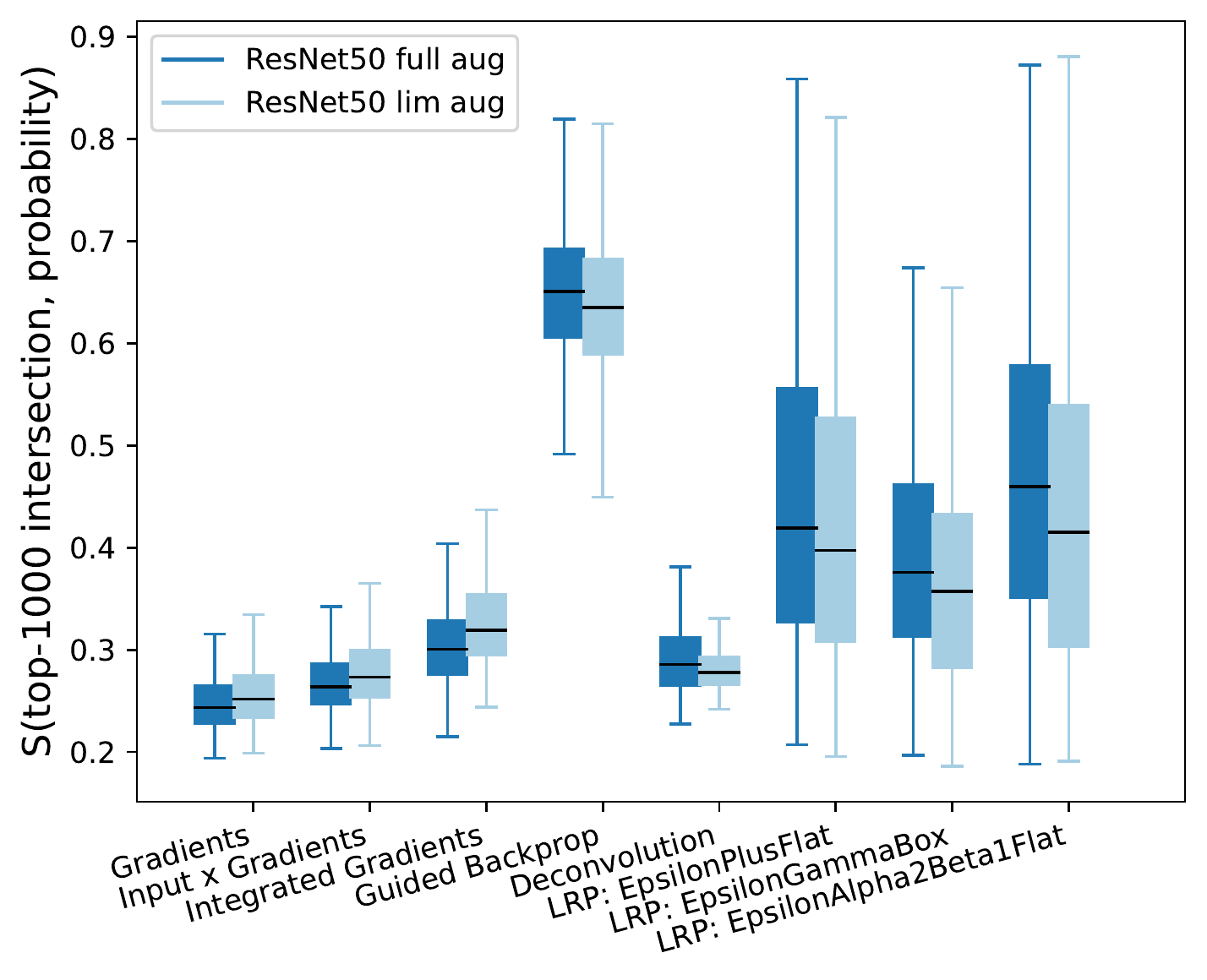}    
   \caption{Translate, $[-0.06, 0.06]$}
   \label{fig:boxplot_ResNet50_Translate_topk}
  \end{subfigure}
\end{figure}

\end{document}